\documentclass{article} 
\usepackage{arxiv}

\usepackage[utf8]{inputenc} 
\usepackage[T1]{fontenc}    
\usepackage{hyperref}       
\usepackage{url}            
\usepackage{booktabs}       
\usepackage{amsfonts}       
\usepackage{nicefrac}       
\usepackage{microtype}      
\usepackage{xcolor}         
\usepackage{graphicx}
\usepackage{amsmath}
\usepackage{subcaption}
\usepackage{amsthm}

\usepackage{natbib}
\usepackage{multirow}
\usepackage{wrapfig}
\usepackage{algorithm}
\usepackage{algpseudocode}
\usepackage{fancyhdr}
\newtheorem{proposition}{Proposition}

\title{How Faithful Is Trajectory-Based Data Attribution? Error Sources, Remedies, and Practical Guidelines}

\author{
Junwei Deng$^{1}$\thanks{Email: \texttt{junweid2@illinois.edu}},
Pingbang Hu$^{1}$,
Suliang Jin$^{2}$,
Hao Lu$^{3}$,
Jiachen T. Wang$^{3}$,
Shichang Zhang$^{4}$,
Jiaqi W. Ma$^{1}$ \\
\\
$^{1}$University of Illinois Urbana-Champaign \\
$^{2}$University of Michigan \\
$^{3}$Princeton University \\
$^{4}$Harvard University
}

\begin{document}

\maketitle

\begin{abstract}
    \emph{Trajectory-based data attribution} methods estimate the influence of training samples on model predictions by unrolling the training trajectory. They are widely used in applications such as data selection, data valuation, and model diagnosis, but there is a lack of comprehensive error analysis of these methods, raising concerns about method faithfulness and hindering reliable deployment. In this work, we provide the first systematic analysis of error sources in trajectory-based data attribution, together with concrete remedies to mitigate them and practical guidelines for downstream use. We organize the total error into three categories, \emph{config-level}, \emph{algorithm-level}, and \emph{system-level}. Building on this taxonomy, we make three contributions. First, we identify optimizer mismatch as the dominant config-level error: existing methods derive their attribution under the assumption of SGD, even for models trained with the modern de facto optimizer AdamW. We propose AdamW-influence to fully account for AdamW's optimization dynamics, yielding improvements from 10\% to over 300\% in Spearman correlation between estimated and ground-truth influence across four settings spanning MLP, CNN, GPT-2, and Llama 3.2-1B. Second, we isolate the remaining algorithm-level error arising from the first-order Taylor approximation, identify the learning rate and trajectory length as factors governing the error magnitude, and derive a closed-form error proxy that can be evaluated along the original trajectory without retraining. Third, we translate these insights into practical guidelines for data selection by unifying offline and online strategies under a $K$-step look-ahead framework. Under this framework, online selection with a short horizon often matches or exceeds offline, and the optimal horizon can be tuned jointly with the learning rate. Together, these results turn the framework into an actionable selection recipe for practitioners.
\end{abstract}

\section{Introduction}
As modern AI systems increasingly rely on large-scale training data, \emph{data attribution},  which estimates the influence of individual training samples on model behavior, has emerged as a central tool for improving performance and explainability~\citep{deng2025survey}. Among existing approaches, \emph{trajectory-based data attribution} (hereafter, \emph{trajectory-based attribution}), such as SGD-influence~\citep{hara2019data}, TracIn~\citep{pruthi2020estimating}, SOURCE~\citep{bae2024training}, and DVEmb~\citep{wang2025capturing}, has emerged as the research frontier of data attribution. These methods relax the strong convexity and convergence assumptions that influence-function-style methods rely on but that rarely hold in modern deep learning~\citep{bae2022if}. Trajectory-based attribution underlies a growing range of applications, e.g., data cleaning~\citep{hara2019data}, data selection~\citep{wang2024greats, pmlr-v235-xia24c, hu2026a}, data valuation~\citep{wang2025data}, and model diagnosis~\citep{zhang2025gets, wang2025capturing}. Recent advances, such as gradient decomposition and random projection~\citep{wang2025data}, have further scaled these methods to modern model sizes, bringing them from a conceptual tool to a practical one.

As trajectory-based attribution matures into a practical tool driving downstream decisions, the reliability of these decisions hinges directly on the accuracy of the attribution scores. Yet practitioners deploying these attribution methods today cannot answer fundamental questions that arise in practice: How large is the attribution error in a training run? Where does the error originate, from a mismatch between the attribution formula and the actual training dynamics, or from approximations inherent in the algorithm itself? And how do routine training choices, such as the optimizer or the learning rate, affect the error? Despite the growing deployment of trajectory-based attribution, no systematic analysis answers these questions, leaving downstream applications exposed to attribution errors of unknown magnitude and origin, hindering broader deployment of these methods in practice.

\begin{figure}[t]
    \centering
    \includegraphics[width=\linewidth]{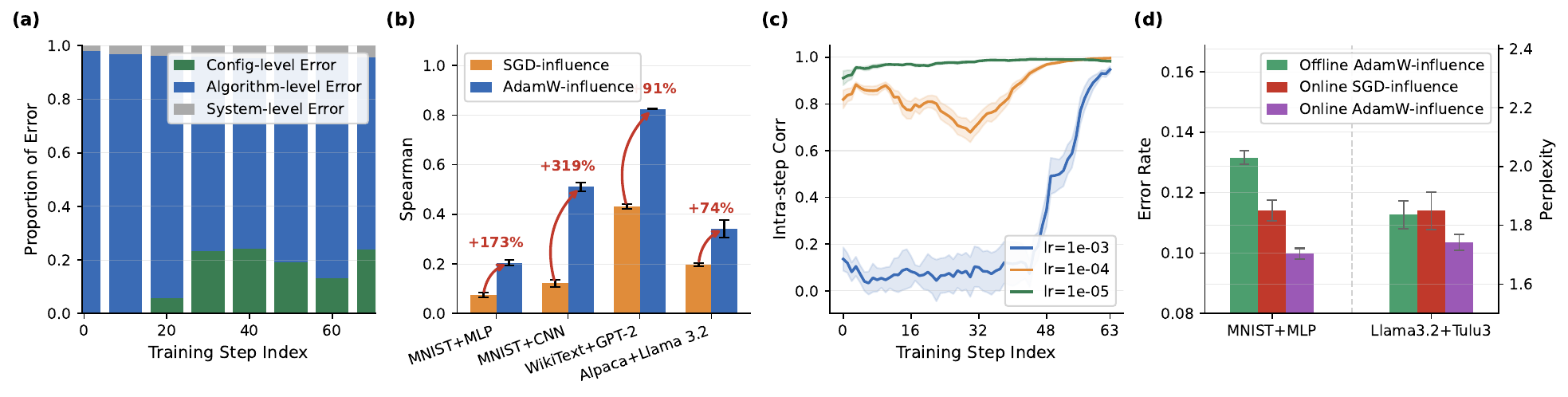}
    \caption{\textbf{An error taxonomy of trajectory-based attribution and our three key contributions.}
\textbf{(a) Three error sources and how their proportion changes along the training trajectory.} The absolute error of SGD-influence on MNIST+MLP decomposes into config-level (green, optimizer mismatch), algorithm-level (blue, first-order Taylor), and system-level (gray) components.
\textbf{(b) The proposed AdamW-influence algorithm corrects the dominant config-level error.} AdamW-influence (blue) improves over SGD-influence (orange) significantly in attribution fidelity measured by the Spearman $\rho$ against TSLOO ground truth across four settings.
\textbf{(c) Dissecting the algorithm-level error shows that learning rate and trajectory length are governing factors.} Intra-step Spearman $\rho$ of AdamW-influence rises as the learning rate decreases and as training steps approach the trajectory end (shorter remaining trajectory).
\textbf{(d) Our error corrections through aligned optimizer and short-horizon attribution can significantly boost data selection quality.} Online AdamW-influence (purple) outperforms online SGD-influence (red) and offline AdamW-influence (green) on MLP image classification (error rate, $\downarrow$) and Llama 3.2-1B SFT on Tulu3 (perplexity, $\downarrow$).}
    \label{fig:first-figure}
    \vspace{-20pt}
\end{figure}

In this work, we provide \textbf{the first systematic error analysis of trajectory-based attribution, with remedies and practical guidelines for its downstream deployment}. We specifically focus on the state-of-the-art efficient version~\citep{wang2025capturing} of SGD-influence~\citep{hara2019data} method for our analysis. We organize the total attribution error into three categories (Figure~\ref{fig:first-figure}a). \emph{Config-level error} arises when the attribution formula does not faithfully reflect the actual training procedure, e.g., optimizer, batch composition, data curriculum or learning rate schedule. \emph{Algorithm-level error} arises from approximations for efficiency, such as the first-order Taylor expansion. \emph{System-level error} that is generally irreducible by algorithm design, such as errors caused by floating-point precision and GPU non-determinism. Building on this taxonomy, we make three contributions: analysis and a new algorithm to address the config-level error, dissection and proxy to remedy the algorithm-level error, and translating these analyses into practical guidelines for an impactful downstream application: data selection.

To address config-level error, we focus on its most impactful source: \emph{optimizer mismatch}. Existing methods~\citep{hara2019data,wang2025capturing,pruthi2020estimating} derive their attribution under the assumption of SGD, yet modern models~\citep{grattafiori2024llama,dosovitskiy2020image} are almost universally trained with AdamW~\citep{loshchilov2017decoupled}, introducing systematic deviation at every attribution step. We propose \textbf{AdamW-influence}, which correctly accounts for AdamW's momentum states and coordinate-wise scaling. Across four settings spanning MLP, CNN, GPT-2, and Llama 3.2-1B, it improves the Spearman correlation between estimated and ground-truth influence by 10\% to over 300\% against the SGD-based baseline (Figure~\ref{fig:first-figure}b). This correction is also the necessary first step before the remaining algorithm-level error can be cleanly isolated. 

With config-level error corrected, we turn to the algorithm-level error from the \emph{first-order Taylor expansion}. We identify \textbf{two factors that govern its magnitude, the learning rate and the trajectory length}, controlling per-step magnitude and accumulation depth, respectively (Figure~\ref{fig:first-figure}c). At the per-sample level, we further derive a closed-form error proxy evaluable along the original trajectory without retraining, which correlates strongly (Spearman $\rho \in [0.83, 0.89]$) with ground-truth error norms from trajectory-specific leave-one-out retraining.

Finally, we translate these insights into \textbf{practical guidelines for data selection}, unifying offline and online strategies under a $K$-step look-ahead framework, where $K$ controls the horizon over which attribution informs selection. Our error analysis directly explains three empirical findings (Figure~\ref{fig:first-figure}d): optimizer alignment transfers from attribution fidelity to selection performance, with online AdamW-influence outperforming its SGD-based baseline; online selection with small to moderate $K$ matches or exceeds offline, since offline corresponds to the longest horizon and accumulates the largest error; and the optimal $K$ grows as the learning rate decreases, since smaller per-step learning rates allow the $K$-step summation to extend further before error dominates. These results turn the error analysis to actionable selection recipe for practitioners.

\section{Error analysis of trajectory-based data attribution}

\subsection{Preliminaries: trajectory-based attribution and SGD-influence}
Trajectory-based data attribution methods evaluate a data point's influence by tracing the exact sequence of mini-batches and the dynamics of the training process, rather than relying only on the final model parameters. Consider an optimization process that runs for $T$ steps, yielding a parameter trajectory $\theta_0, \theta_1, \dots, \theta_{T}$. Let $z^*$ be a training data point in mini-batch $\mathcal{B}_{t^*}$ at iteration $t^*$. The exact counterfactual impact of removing $z^*$ from the parameter update at step $t^*$ on the loss at a validation point $z^{(\text{val})}$ is the Trajectory-Specific Leave-One-Out (TSLOO) error, originally conceptualized by \citet{hara2019data}:
\begin{equation}\label{eq:tsloo}
    \text{TSLOO}(z^*; z^{(\text{val})}) := \ell(\theta'_{T,z^*}, z^{(\text{val})}) - \ell(\theta_T, z^{(\text{val})}),
\end{equation}
where $\theta'_{T,z^*} \in \mathbb{R}^p$ is the final model parameters obtained if $z^*$ had been excluded from the parameter update at step $t^*$, and $p$ is the parameter size.

\citet{hara2019data} approximate the TSLOO error via a first-order Taylor expansion of the parameter trajectory in a perturbation scalar $\epsilon$ injected at step $t^*$, yielding the \emph{SGD-influence}:
\begin{equation}\label{eq:sgd-influence}
    \ell(\theta'_{T,z^*}, z^{(\text{val})}) - \ell(\theta_T, z^{(\text{val})}) \approx \eta_{t^*}\, \nabla_\theta \ell(\theta_T, z^{(\text{val})})^\top \left[ \prod_{t=t^*+1}^{T-1} (I - \eta_t H_t) \right] \nabla_\theta \ell(\theta_{t^*}, z^*),
\end{equation}
where $\eta_t$ is the learning rate and $H_t = \nabla_\theta^2 \ell(\theta_t, \mathcal{B}_t)$ is the per-step Hessian.
\citet{wang2025capturing} reduce the cost to $\mathcal{O}(T)$ under the GGN approximation $H_t \approx \frac{1}{|\mathcal{B}_t|}\sum_{z\in\mathcal{B}_t} g_{t,z} g_{t,z}^\top$ (with $g_{t,z}=\nabla_\theta \ell(\theta_t, z)$) by extracting test-independent components into per-sample embeddings, computed via a backward recurrence over a summary matrix $W^{(t)} \in \mathbb{R}^{p\times p}$. We refer to this efficient implementation as SGD-influence throughout this paper and defer a detailed discussion of \cite{wang2025capturing}'s implementation to Appendix~\ref{app:sgd-influence}.

\subsection{The error sources of trajectory-based attribution}\label{sec:error-type}

The \emph{attribution error} is the discrepancy between the estimated attribution score and the ground-truth TSLOO (Eq.~\ref{eq:tsloo}). We organize its sources along the trajectory into three categories: config-level, algorithm-level, and system-level.

\textbf{Config-level error.}
Config-level error arises whenever the attribution formula does not faithfully reflect a component of the actual training procedure. Potential sources include the optimizer, batch composition, data curriculum, and learning rate schedule. Among these, batch composition, data curriculum and learning rate schedule are aligned straightforwardly by replaying the recorded training log, and SGD-influence already incorporates both. \emph{Optimizer mismatch}, by contrast, is neither addressed by existing methods nor trivial to correct: current methods derive their attribution under the assumption of SGD~\citep{hara2019data,wang2025capturing}, even for models trained with modern de facto optimizer AdamW. In Section~\ref{sec:config-level-error}, we address this dominant config-level error and show that the correction will significantly improve attribution fidelity.

\textbf{Algorithm-level error.}
Even when the attribution formula perfectly matches the training dynamics, computing the exact TSLOO is intractable for nonlinear models, and trajectory-based attribution relies on approximations such as a first-order Taylor expansion of the parameter trajectory. Algorithm-level error is the gap introduced by these approximations, and its profile depends on the specific approximation strategy. We analyze it in the context of AdamW-influence, whose attribution relies on a first-order Taylor expansion. In Section~\ref{sec:approx-error}, we dissect this error, identify the factors governing its magnitude, and derive a closed-form proxy for per-sample analysis.

\textbf{System-level error.}
System-level errors, such as floating-point precision and GPU non-determinism, are not specific to attribution methods, and even compute the TSLOO by directly retraining the model also suffer from this type of error. Therefore, they are largely irreducible by attribution algorithm design. Empirically, we find this irreducible part to be relatively small compared to the total error. We thus focus on analyzing and reducing the other two types of errors.

\section{Correcting the dominant config-level error: AdamW-influence}\label{sec:config-level-error}
As established in Section~\ref{sec:error-type}, optimizer mismatch is the dominant source of config-level error and requires non-trivial solution. In Section~\ref{sec:adamw-dvemb}, we derive AdamW-influence, which extends the SGD-influence to correctly account for the modern de facto optimizer AdamW~\citep{loshchilov2017decoupled}.
We then evaluate its attribution fidelity in Section~\ref{sec:fidelity}, demonstrating substantial improvements across four settings.

\subsection{AdamW-influence}\label{sec:adamw-dvemb}

Both SGD-influence and AdamW-influence estimate the same TSLOO target (Eq.~\ref{eq:tsloo}) but unroll different optimizers along the trajectory. SGD-influence unrolls the SGD updates for simplicity, even when the trajectory is actually from a different optimizer such as AdamW. Our AdamW-influence performs the analogous estimate but unrolls the actual AdamW updates.

\textbf{Challenges for unrolling the AdamW updates.} Two features distinguish AdamW from SGD and make the AdamW unrolling non-trivial: (i) momentum and second-moment states that carry perturbations across steps, and (ii) a coordinate-wise adaptive scaling that makes the update nonlinear in the gradient. The SGD-influence derivation hinges on the update being linear in the per-step gradient, so that injecting a parameter-level perturbation is equivalent to removing $z^*$ from the batch. Under AdamW this equivalence breaks: the second moment depends quadratically on the gradient and enters the update nonlinearly through the denominator, so a parameter-level perturbation no longer corresponds to removing $z^*$ from the batch.

\textbf{AdamW unrolling.} To address this, AdamW-influence injects the perturbation at the gradient level rather than the parameter level, and propagates it through the joint dynamics of the parameters and the two moment states.

\begin{proposition}[AdamW-influence, informal]\label{prop:adamw-influence}
The AdamW-influence score for every training sample can be computed in $\mathcal{O}(T)$ via a backward recurrence over the training trajectory, matching the asymptotic complexity of SGD-influence.
\end{proposition}

The proof extends the backward computation of SGD-influence, with the main challenge and innovation being the propagation of the gradient-level perturbation through AdamW's coupled moment states and adaptive scaling. The full derivation, transition operators, and per-sample recurrence are deferred to Appendix~\ref{app:adamw-derivation}, and the cost analysis to Appendix~\ref{app:efficient-computation}.

\subsection{Fidelity evaluation}\label{sec:fidelity}

We evaluate the attribution fidelity of AdamW-influence against the SGD-influence baseline by measuring the Spearman rank correlation between predicted attribution scores and ground-truth TSLOO scores (Eq.~\ref{eq:tsloo}) obtained via retraining.
Results in Table~\ref{tab:fidelity-adamw} show that AdamW-influence outperforms SGD-influence in four settings: an MLP and a CNN trained on MNIST, GPT-2 continually pretrained on WikiText-2, and Llama~3.2-1B fine-tuned on Alpaca. The improvements range from 10\% to over 300\%. 
For each setting, we also vary the learning rate, which we find to be an important factor in the error and study systematically in Section~\ref{sec:approx-error}. Full training details and hyperparameters are provided in Appendix~\ref{app:exp-details}.

\begin{table}[h]
\centering
\caption{Fidelity comparison (Spearman $\rho$ against TSLOO ground truth) between AdamW-influence and SGD-influence when training with \textbf{AdamW}. Best result per row in \textbf{bold}. $\Delta\%$ is the relative improvement of AdamW-influence over SGD-influence. Full results across all data sizes in Appendix~\ref{app:additional-fidelity}.}
\resizebox{0.75\textwidth}{!}{%
\begin{tabular}{llccc}
\toprule
Setting & LR & AdamW-influence & SGD-influence & $\Delta\%$ \\
\midrule
\multirow{3}{*}{MNIST+MLP}        & 1e-3 & $\mathbf{0.205{\pm}0.011}$ & $0.075{\pm}0.009$ & $+173\%$ \\
                                  & 1e-4 & $\mathbf{0.294{\pm}0.014}$ & $0.242{\pm}0.016$ & $+21\%$ \\
                                  & 1e-5 & $\mathbf{0.786{\pm}0.013}$ & $0.715{\pm}0.012$ & $+10\%$ \\
\cmidrule(lr){1-5}
\multirow{3}{*}{MNIST+CNN}        & 1e-3 & $\mathbf{0.090{\pm}0.008}$ & $0.015{\pm}0.009$ & $+500\%$ \\
                                  & 1e-4 & $\mathbf{0.511{\pm}0.017}$ & $0.122{\pm}0.015$ & $+319\%$ \\
                                  & 1e-5 & $\mathbf{0.791{\pm}0.007}$ & $0.526{\pm}0.007$ & $+50\%$ \\
\cmidrule(lr){1-5}
\multirow{3}{*}{WikiText+GPT-2}   & 1e-4 & $\mathbf{0.734{\pm}0.003}$ & $0.372{\pm}0.006$ & $+97\%$ \\
                                  & 5e-5 & $\mathbf{0.825{\pm}0.002}$ & $0.432{\pm}0.009$ & $+91\%$ \\
                                  & 1e-5 & $\mathbf{0.842{\pm}0.006}$ & $0.480{\pm}0.015$ & $+75\%$ \\
\cmidrule(lr){1-5}
\multirow{3}{*}{Alpaca+Llama~3.2} & 6e-6 & $\mathbf{0.276{\pm}0.047}$ & $0.156{\pm}0.019$ & $+77\%$ \\
                                  & 2e-6 & $\mathbf{0.342{\pm}0.036}$ & $0.197{\pm}0.007$ & $+74\%$ \\
                                  & 5e-7 & $\mathbf{0.665{\pm}0.013}$ & $0.330{\pm}0.017$ & $+102\%$ \\
\bottomrule
\end{tabular}%
}
\vspace{-10pt}
\label{tab:fidelity-adamw}
\end{table}


\section{Dissecting the algorithm-level error }\label{sec:approx-error}

Section~\ref{sec:config-level-error} corrected the dominant config-level error by aligning the attribution formula with the AdamW dynamics. We now turn to the algorithm-level error, which arises from the first-order Taylor expansion that AdamW-influence\footnote{And other trajectory-based attribution methods~\citep{hara2019data,wang2025capturing} more broadly.} uses to make attribution tractable. This section dissects the algorithm-level error in three steps. Section~\ref{ssec:error-accumulation} empirically locates its dominant error source, isolating the \emph{update-estimation error} as the bottleneck. Section~\ref{sec:two-factors} identifies the two factors, learning rate and trajectory length, that jointly govern its magnitude. Section~\ref{sec:error-proxy} derives a closed-form error proxy that estimates per-sample error along the original training trajectory without retraining.

\subsection{Dominant error from estimating the parameter update}\label{ssec:error-accumulation}

Recall from Section~\ref{sec:error-type} that the algorithm-level error of AdamW-influence decomposes into the \emph{update-estimation error}, the gap between the true parameter update $\Delta\theta_{z^*} = \theta'_{T,z^*} - \theta_{T}$ and estimated parameter update $\text{AdamW-influence}_{z^*}$, and a \emph{residual} collecting higher-order terms and system-level error. To locate which component dominates, and to quantify how much of the total attribution error was previously absorbed by optimizer mismatch, we decompose the absolute SGD-influence error of a sample $z^*$ into three additive components:
\begin{equation}
\resizebox{0.92\linewidth}{!}{%
    $\begin{aligned}
    \big|\mathrm{Error_{\text{SGD}}}_{z^*}\big| \;=\; & \underbrace{\big|\mathrm{Error_{\text{SGD}}}_{z^*}\big| - \big|\mathrm{Error_{\text{AdamW}}}_{z^*}\big|}_{\text{{\color{green!50!black} Green}: optimizer mismatch}} \;+\; \underbrace{\big|\nabla\ell(\theta_T,z^{(\text{val})}) \cdot (\Delta\theta_z^* - \text{AdamW-influence}_{z^*})\big|}_{\text{{\color{blue} Blue}: update-estimation error}} \;+\; \underbrace{\text{rest}}_{\text{{\color{gray} Grey}: residual}},
    \end{aligned}$%
}
\end{equation}
$\big|\mathrm{Error}_{\text{SGD}z^*}\big|$ and $\big|\mathrm{Error}_{\text{AdamW}z^*}\big|$ are the absolute errors of influence estimate by SGD-influnce and AdamW-influence with the TSLOO ground-truth, respectively. The green term measures the error share corrected by AdamW-influence; the blue term is the update-estimation error remaining after optimizer alignment; the grey term collects all higher-order remainders and system-level error.

\begin{figure}[h]
    \centering
    \includegraphics[width=\linewidth]{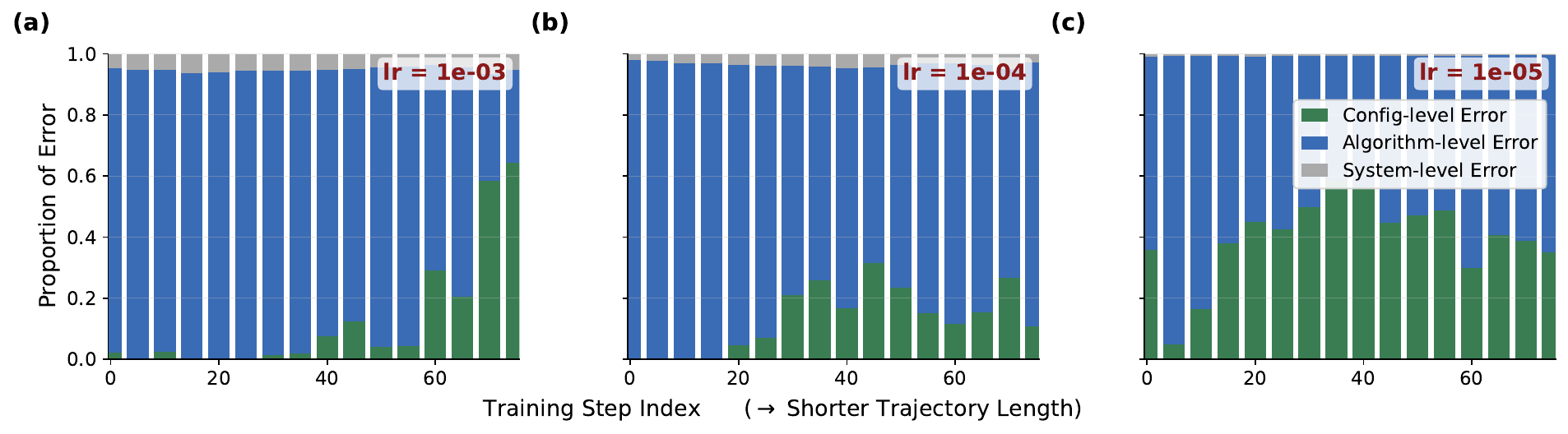}
    \caption{\textbf{Decomposition of $\big|\mathrm{Error_{\text{SGD}}}\big|$ across the training trajectory of an MLP trained on MNIST.} Panels (a)--(c) sweep three learning rates: \textbf{(a)} $\eta=10^{-3}$, \textbf{(b)} $\eta=10^{-4}$, \textbf{(c)} $\eta=10^{-5}$. The absolute SGD-influence error is decomposed into three additive components, where each category is represented by its dominant source: {\color{green!50!black} \textbf{Green}}, config-level error from optimizer mismatch, the share fixed by switching to AdamW-influence; {\color{blue} \textbf{Blue}}, algorithm-level error from the first-order Taylor expansion (the update-estimation error remaining after optimizer alignment); and {\color{gray} \textbf{Grey}}, residual error absorbing all higher-order remainders together with system-level effects (e.g., floating-point precision, GPU non-determinism). Errors are aggregated into bins of 5 training steps; larger step indices correspond to shorter remaining trajectory length $T-t^*$.}
    \label{fig:error-decomposition}
    \vspace{-15pt}
\end{figure}

Figure~\ref{fig:error-decomposition} reports this decomposition on an MLP trained on MNIST across three learning rates ($10^{-3}$, $10^{-4}$, $10^{-5}$), with errors aggregated into bins of 5 training steps (full setup in Appendix~\ref{app:exp-details}). Two patterns stand out. (1) Once optimizer mismatch is corrected, \textbf{the update-estimation error accounts for the overwhelming majority} of the remaining error, while the higher-order residual is consistently small. We thus focus on analyzing and correcting the dominant update-estimation error for the rest of this section. (2) The optimizer-mismatch share grows as the learning rate decreases and as training progresses, indicating that \textbf{optimizer alignment becomes \emph{more} important in low learning rate or short trajectory length cases}---precisely the regimes most relevant to data selection in our later analysis.

\subsection{Two factors governing the algorithm-level error}
\label{sec:two-factors}

Having established that the update-estimation error dominates the algorithm-level error, we investigate what governs its magnitude. The error has a intuitively simple structure: at each step, a first-order Taylor expansion linearizes the parameter update around the unperturbed trajectory, incurring a per-step error whose magnitude scales with the size of that step; these per-step errors then accumulate as the perturbation propagates to the end of training. This structure points to two controllable training quantities: the \emph{learning rate}, which sets the magnitude of each per-step parameter update and thus the per-step linearization error; and the \emph{trajectory length} $T-t^*$, the distance from the injection step $t^*$ to the trajectory terminal $T$, which determines how many per-step errors are summed into the final attribution error.

To verify the influence of each factor, we sweep three learning rates on an MLP and a CNN trained on MNIST, evaluating attribution quality across the trajectory through two metrics: the \emph{error norm} $\|\Delta\theta_{z^*} - \text{AdamW-influence}_{z^*}\|^2$, which directly measures parameter-space error (lower is better); and the \emph{intra-step Spearman correlation} between attributed and ground-truth per-sample influences within each training step, which reflects the ability to rank samples within a batch (higher is better). Since the trajectory length $T-t^*$ is itself indexed by the training step, a single sweep over training steps traces both factors at once, with each curve isolating the learning-rate effect at a fixed trajectory length.

\begin{figure}[h]
    \centering
    \includegraphics[width=\linewidth]{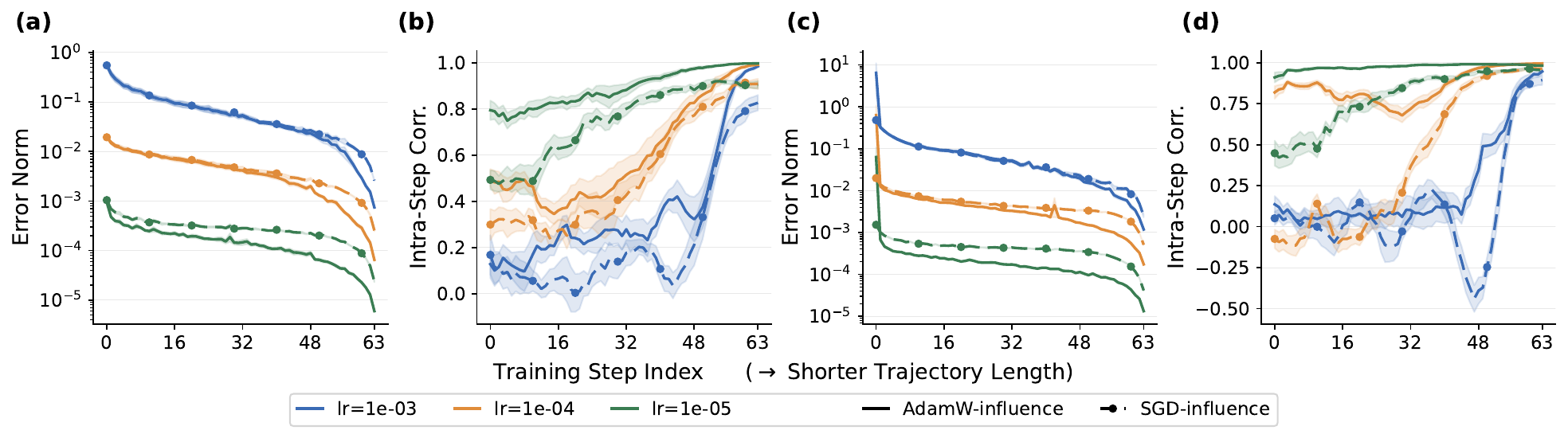}
    \caption{\textbf{Error norm and intra-step Spearman correlation across three learning rates}, on \textbf{(a, b)} an MLP and \textbf{(c, d)} a CNN trained on MNIST. Panels (a, c) report error norm ($\downarrow$); panels (b, d) report intra-step Spearman correlation ($\uparrow$, smoothed over 5 consecutive steps). The training step index runs from $0$ to $T-1$; larger indices correspond to \emph{shorter} trajectory lengths $|T - t^*|$, since the perturbation has fewer remaining steps to propagate. AdamW-influence (solid) and SGD-influence (dashed) are shown for reference.}
    \label{fig:two-factors}
\end{figure}

\textbf{Learning rate.}
\label{sec:factor-lr}
Larger learning rates push per-step parameter updates further from the linearization point, amplifying the higher-order terms that the first-order approximation discards. Figure~\ref{fig:two-factors} confirms this dependence across architectures and metrics: the error norm increases monotonically with the learning rate throughout the trajectory, and the intra-step correlation degrades correspondingly, with $\eta=10^{-5}$ ($\eta=10^{-3}$) achieving the highest (lowest) correlation throughout training.

\textbf{Trajectory length.}
\label{sec:factor-trace}
Independently of the magnitude of each step, errors propagate and accumulate along the training trajectory. In Figure~\ref{fig:two-factors}, we see that as the trajectory length $T - t^*$ shrinks (i.e., as the training step index approaches $T$), the error norm decreases and the intra-step correlation increases across all three learning rates. \textbf{Notably, AdamW-influence reaches near-perfect intra-step correlation at very short trajectory lengths for all three learning rates}, indicating that short-horizon attribution is reliable even at learning rates where full-trajectory attribution is noisy. This compounding behavior between the two factors will be central to the practical implications in Section~\ref{sec:practical-data-sel}.

\subsection{Error proxy}
\label{sec:error-proxy}

Beyond identifying trajectory-level factors that govern attribution error magnitude, we also try to identify per-sample errors to facilitate applications like data selection. Computing per-sample error directly requires TSLOO retraining that is computationally prohibitive. We thus derive a closed-form \emph{error proxy} that can be evaluated along the original training trajectory.

\begin{proposition}[Error proxy for update-estimation error]
\label{prop:proxy}
Under the AdamW training dynamics in Section~\ref{sec:adamw-dvemb}, 
the update-estimation error at the final step admits a closed-form 
expression as the sum of per-step residuals $r_t \in \mathbb{R}^p$ 
propagated through the linearized dynamics, where per-coordinate 
residual $r_{t,i}$ is
\begin{equation}\label{eq:proxy-unroll}
\bigl|r_{t,i}\bigr| \;\sim\; \eta_t
\left(\frac{\|\dot\theta_t\|^2}{\sqrt{\hat v_{t,i}}}
+ \frac{[H_t\dot\theta_t]_i^{\,2}}{\hat v_{t,i}}\right).
\end{equation}
Here $\dot\theta_t$ denotes the parameter perturbation propagated by 
AdamW-influence (the derivative of the parameter trajectory with 
respect to the injected perturbation $\epsilon$ at $\epsilon = 0$), 
and $\hat v_{t,i}$ is AdamW's bias-corrected second-moment estimate 
at coordinate $i$. The full unrolled form and derivation are deferred 
to Appendix~\ref{app:error-proxy-derivation}.
\end{proposition}

Equation~\eqref{eq:proxy-unroll} reflects the two error governing factors in Section~\ref{sec:two-factors}. Per-step scaling by $\eta_t$ and accumulation over trajectory length, and additionally exposes a sample-distinguishing term $\|\dot\theta_t\|^2 + [H_t\dot\theta_t]_i^{2}$ that captures training-stage effects unique to each sample. 

\begin{figure}[h]
    \centering
    \includegraphics[width=\linewidth]{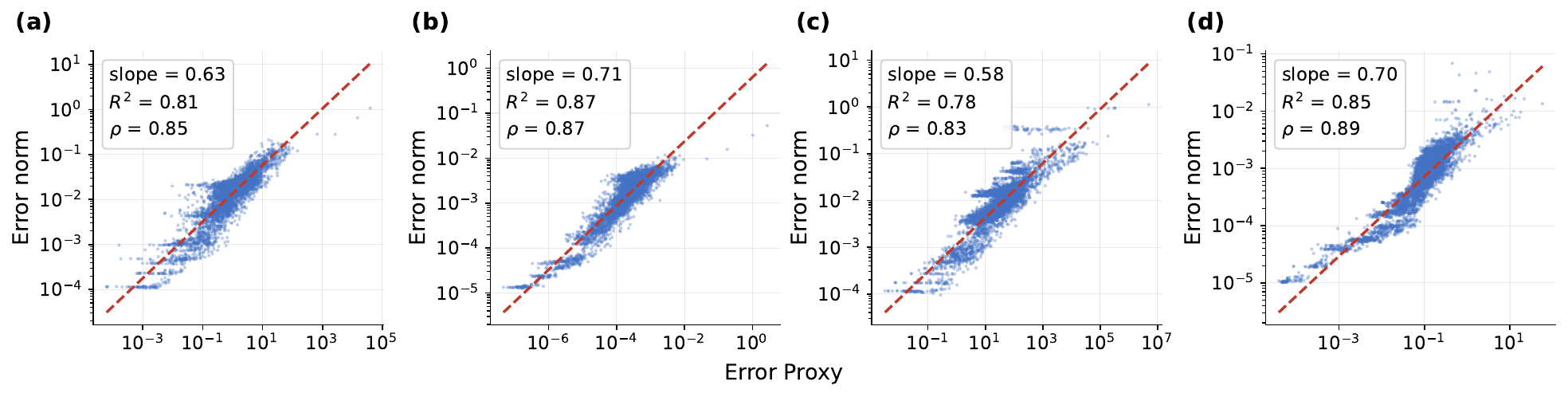}
    \caption{\textbf{The error proxy tracks the ground-truth error norm $\|\Delta\theta - \text{AdamW-influence}\|$ across training samples.} Each point is a single training sample; axes are on log-log scale. Panels: \textbf{(a)} MLP, $\eta=10^{-3}$; \textbf{(b)} MLP, $\eta=10^{-4}$; \textbf{(c)} CNN, $\eta=10^{-3}$; \textbf{(d)} CNN, $\eta=10^{-4}$. Both architectures are trained on MNIST. The red dashed line is a linear fit; insets report its slope and $R^2$, along with the Spearman rank correlation $\rho$ between proxy and ground truth.}
    \label{fig:proxy-validation}
    \vspace{-15pt}
\end{figure}

We validate the proxy against ground-truth error norms on an MLP and a CNN trained on MNIST, each at two learning rates. Figure~\ref{fig:proxy-validation} shows Spearman rank correlation $\rho \in [0.83, 0.89]$ across all four settings.
Crucially, the proxy reuses quantities already maintained by AdamW-influence plus one Hessian-vector product per step, keeping total cost at $\mathcal{O}(T)$. It makes the proxy a practical signal for trajectory-aware decisions downstream, including the error-driven horizon analysis in Section~\ref{sec:practical-data-sel}.

\section{Practical implications in data selection}\label{sec:practical-data-sel}

Sections~\ref{sec:config-level-error} and~\ref{sec:approx-error} together provide a systematic understanding of where the error in trajectory-based attribution originates and how to mitigate it. We now translate these insights into practical guidelines for data selection, one of the most consequential downstream applications of trajectory-based attribution. Specifically, we use the error analysis to predict which selection strategies should work better and verify these predictions empirically.

\subsection{A unified $K$-Step look-ahead framework for data selection}
\label{sec:unified-K}

\begin{figure}[t]
    \centering
    \includegraphics[width=\linewidth]{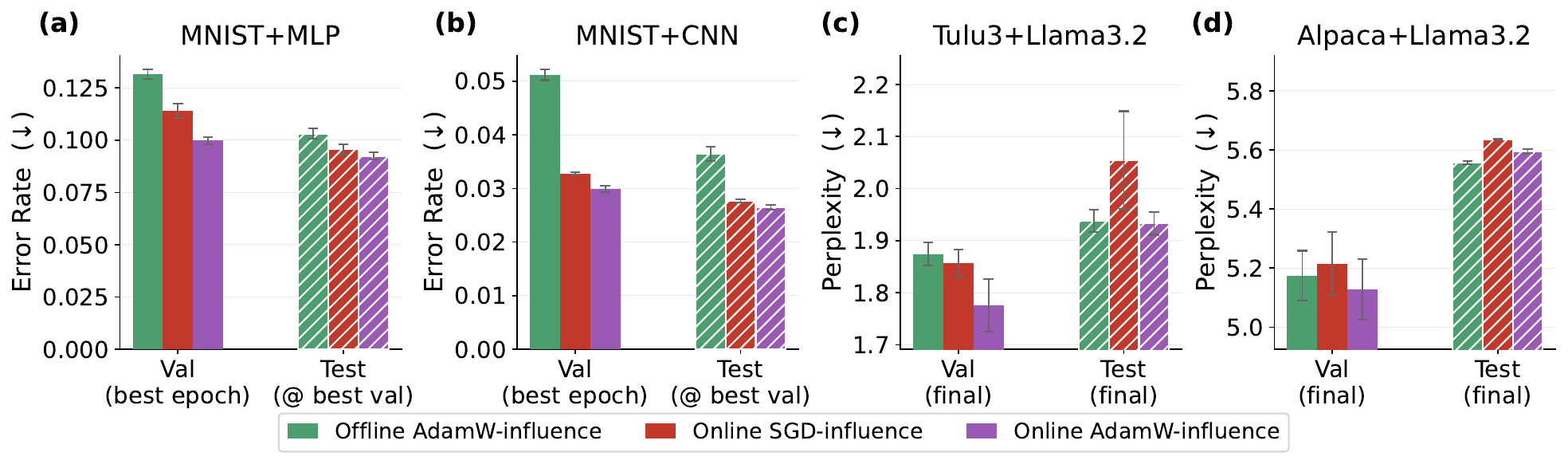}
    \caption{\textbf{Online AdamW-influence consistently outperforms online SGD-influence, and matches or exceeds offline AdamW-influence across four settings.} \textbf{(a)} an MLP and \textbf{(b)} a CNN trained on MNIST (error rate, $\downarrow$); Also, Llama 3.2-1B fine-tuned on \textbf{(c)} Tulu3 and \textbf{(d)} Alpaca (perplexity, $\downarrow$). For MNIST settings, ``Val'' is at the best epoch and ``Test'' is reported at that epoch; for Llama 3.2-1B settings, both are at the final step. $K=10$ for MLP, $K=5$ for CNN, and $K=2$ for Llama 3.2-1B.}
    \label{fig:offline-adam-sgd-datasel}
    \vspace{-15pt}
\end{figure}

Trajectory-based attribution can be used for data selection in two ways: \emph{offline}, which runs a full training pass, computes attribution against a held-out validation set using the entire trajectory, and retrains from scratch on the selected subset; and \emph{online}, which selects samples on-the-fly during a single training run using a short-horizon attribution score evaluated at the current parameters~\cite{deng2025survey}. Despite their procedural differences, we show that they can be conceptually unified along a single axis: the look-ahead horizon of each selection decision.

\textbf{Unified $K$-Step Setup.} The horizon $K$ controls how far ahead each candidate is scored. At step $t$, instead of asking ``does including $z$ help validation loss right now?'' we ask ``does including $z$ help validation loss $K$ steps from now?''. Larger $K$ looks further into the future and captures more downstream effect; smaller $K$ stays close to the present.

Concretely, at each step $t$ we draw a random candidate batch $\mathcal{C}_t$ of size $N$, score each candidate $z \in \mathcal{C}_t$ by $s_t^{(K)}(z) = \nabla_\theta \ell(\theta_t, z^{(\text{val})})^\top \dot{\theta}_{t+K+1}(z)$, and retain the top-$B$ as $\mathcal{B}_t$ for the parameter update. Here $\dot{\theta}_{t+K+1}(z)$ is the parameter perturbation $K$ steps after injecting $z$ at step $t$, computed via the AdamW-influence recurrence of Section~\ref{sec:config-level-error}. Algorithm~\ref{alg:kstep} (Appendix~\ref{app:kstep-pseudocode}) gives the pseudo-code.

This single knob recovers existing strategies as boundary cases: $K = 0$ is the per-step scoring rule used by GREATS~\cite{wang2024greats} (without its greedy selection iteration), and $K = T - t$ is full-trajectory propagation, intractable to run online but approximated by offline selection on a completed reference trajectory. Online selection with finite $K$ thus scans the intermediate horizons between these two extremes. This exposes $K$ directly to the error analysis of Section~\ref{sec:approx-error}: larger $K$ accumulates more approximation error, while smaller $K$ discards information about downstream dynamics.

\subsection{Practical implications and error-analysis explanations}
\label{sec:practical implications}

We experiment on four settings with optimizer AdamW, comparing ``offline AdamW-influence'', ``online SGD-influence'', and ``online AdamW-influence'' with varying $K$. Full setup in Appendix~\ref{app:data-selection-settings}.

\begin{wrapfigure}[22]{r}{0.45\linewidth}
    \centering
    \includegraphics[width=\linewidth]{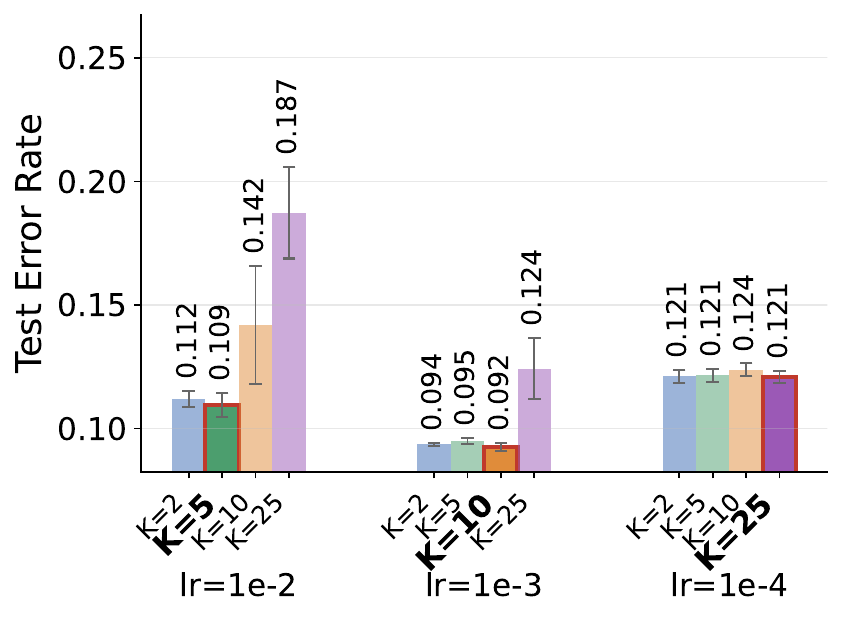}
    \caption{The optimal $K$ increases as the learning rate decreases. We sweep $K \in \{2, 5, 10, 25\}$ across three learning rates using online AdamW-influence for MLP on MNIST. Bars show test error rate at the best validation epoch; the best $K$ per learning rate is highlighted in red.}
    \label{fig:selection-k-sweep}
\end{wrapfigure}

\textbf{Optimizer alignment transfers to downstream selection.} Online AdamW-influence outperforms online SGD-influence in all settings (Figure~\ref{fig:offline-adam-sgd-datasel}), mirroring the attribution-fidelity gains of Section~\ref{sec:fidelity}. The dominant config-level error identified in Section~\ref{sec:error-type} thus propagates directly into selection quality: a selection rule built on a mis-specified optimizer undermines selection performance.

\textbf{Horizon $K$ trades off lookahead information against attribution error.} Larger $K$ collects more information about a candidate's future effect but accumulates more first-order approximation error per Section~\ref{sec:two-factors}; smaller $K$ keeps error low but reflects only short-range effects. We observe this trade-off between online and offline results: in all four settings, online AdamW-influence with a small-to-moderate $K$ matches or exceeds offline AdamW-influence (Figure~\ref{fig:offline-adam-sgd-datasel}), consistent with offline corresponding to the longest horizon $K = T - t - 1$ and accumulating the largest error. We do not claim offline is universally inferior. When we have small learning rate, short training, or late-stage fine-tuning, offline can remain competitive or superior. Within the practical training regimes considered here, online with a small-to-moderate $K$ is a reliable choice.

\textbf{The optimal $K$ shifts with the learning rate.} Sweeping $K \in \{2, 5, 10, 25\}$ across three learning rates on MNIST (Figure~\ref{fig:selection-k-sweep}), the best $K$ is $5$ at $\eta=10^{-2}$, $10$ at $\eta=10^{-3}$, and $25$ at $\eta=10^{-4}$: the optimal horizon grows as the learning rate shrinks. This is precisely what Section~\ref{sec:two-factors} predicts: larger $K$ collects more information about how a candidate influences future training but accumulates more approximation error, and the optimal $K$ balances these two effects. A practical guidance is that $K$ should be tuned jointly with the learning rate rather than fixed across runs; a larger $K$ is not necessarily better.

\section{Related work}\label{sec:related}

\textbf{Trajectory-based data attribution.} Convergence-based methods, influence functions~\citep{koh2017understanding,grosse2023studying}, TRAK~\citep{pmlr-v202-park23c}, and LoGra~\citep{choe2026what}, rely on convergence or permutation-invariance assumptions violated by modern stochastic training~\citep{wang2025capturing}. Trajectory-based attribution methods mitigate these by attributing along the optimization path: SGD-influence~\citep{hara2019data}, TracIn~\citep{pruthi2020estimating}, SOURCE~\citep{bae2024training}, and DVEmb~\citep{wang2025capturing}. Recent extensions handle cross-epoch compounding under SGD~\citep{shi2025accumulative} or stage-level effects with momentum and weight decay~\citep{zhang2025gets}, but none cover AdamW's second-moment coupling at the per-sample level, and all are proposed as standalone methods without comprehensive error analysis. We extend trajectory-based attribution to AdamW and provide the first error taxonomy with a careful analysis.

\textbf{Data attribution for data selection.} \emph{Offline selection} retrains on a subset chosen using attribution from a completed reference trajectory; representative work includes LESS~\citep{pmlr-v235-xia24c}, which adapts \emph{single-step} influence to Adam. \emph{Online selection} chooses samples on-the-fly: GREATS~\citep{wang2024greats} uses a Taylor approximation of the immediate validation effect. Beyond unrolling-based attribution, MATES~\citep{yu2024mates} fits a small data-influence model to oracle scores collected by locally probing the pretraining model, sidestepping first-order approximation entirely at the cost of training an auxiliary model. We unify both regimes into a unified framework.
\section{Conclusion}\label{sec:conclusion}

We presented the first systematic error analysis of trajectory-based data attribution. By organizing the error into a three-level taxonomy, we develop AdamW-influence to remedy the dominant config-level error, identify the learning rate and trajectory length as the two factors governing the algorithm-level error, and derive a closed-form proxy for per-sample algorithm-level error. Building on this analysis, we further provide actionable guidelines for data selection through a unified $K$-step look-ahead framework. We see this as a step toward treating attribution methods as diagnosable estimators rather than black-box scoring functions. The taxonomy and the error analysis provide a foundation for trajectory-based attribution to scale to larger models and be deployed for broader applications. 

\textbf{Limitations.} Our analysis is restricted to trajectory-based attribution methods that use first-order Taylor since these methods represent the state-of-the-art, but extending the error taxonomy to other attribution methods remains open, such as MAGIC~\citep{ilyas2025magic} and Simfluence~\citep{guu2023simfluence}. Our LLM experiments are limited to models up to Llama 3.2-1B due to computation constraints, and validation at larger scales is left to future work. Our empirical evaluation focuses on supervised classification and language modeling; whether the same error patterns and practical guidelines transfer to other regimes such as reinforcement learning or diffusion models is an important direction for future investigation.


\clearpage
\appendix
\section{Detailed review of SGD-influence}\label{app:sgd-influence}

For SGD-influence, \citet{wang2025capturing} reduce the cost to $\mathcal{O}(T)$ under the GGN approximation $H_t \approx \frac{1}{|\mathcal{B}_t|}\sum_{z \in \mathcal{B}_t} g_{t,z} g_{t,z}^\top$ by extracting test-independent components into per-sample embeddings, computed via a backward recurrence over a summary matrix $W^{(t)} \in \mathbb{R}^{p\times p}$ that encodes how a perturbation injected at step $t$ propagates through all subsequent training steps to the final parameters:
\begin{align}
    \text{SGD-influence}_z &= \eta_t (I - W^{(t)}) g_{t,z}, \quad \forall z \in \mathcal{B}_t,\\
    W^{(t-1)} &= W^{(t)} + \frac{1}{|\mathcal{B}_t|} \sum_{z \in \mathcal{B}_t} \text{SGD-influence}_z\, g_{t,z}^\top, 
\end{align}
with $g_{t,z} = \nabla_\theta l(\theta_t, z)$ and $W^{(T-1)} = \mathbf{0}$. We refer to this efficient implementation as SGD-influence throughout this paper.

\section{Derivation of AdamW-influence} \label{app:adamw-derivation}

This appendix gives the complete derivation of AdamW-influence. Since Section~\ref{sec:adamw-dvemb} states only an informal version of Proposition~\ref{prop:adamw-influence}, we begin by establishing the full notation used in this appendix.

\subsection{Setup and notation}\label{app:adamw-notation}

\textbf{AdamW updates.} Consider a data point $z^*$ that first participates in training at iteration $t^*$. The model is trained with AdamW: given the batch gradient $g_t = \nabla_\theta \ell(\theta_t, \mathcal{B}_t)$, the optimizer maintains first and second moment estimates
\[
  m_t = \beta_1 m_{t-1} + (1-\beta_1) g_t,
  \qquad
  v_t = \beta_2 v_{t-1} + (1-\beta_2) g_t^2,
\]
with bias-corrected versions $\hat m_t = m_t/(1-\beta_1^{t+1})$ and $\hat v_t = v_t/(1-\beta_2^{t+1})$. The parameter update is
\[
  \theta_{t+1} = (1-\eta_t\lambda)\,\theta_t - \eta_t\,\frac{\hat m_t}{\sqrt{\hat v_t}+\varepsilon}.
\]
We write $\theta_{t+1} = (1-\eta_t\lambda)\,\theta_t - U_t$, where $U_t = \eta_t \hat m_t / (\sqrt{\hat v_t}+\varepsilon)$, $\lambda$ is the weight decay coefficient, and $\varepsilon$ is the denominator smoothing term.

\textbf{Gradient-level perturbation.} As motivated in Section~\ref{sec:adamw-dvemb}, AdamW's update is nonlinear in the gradient through the second moment, so a parameter-level $\epsilon$-injection (which is equivalent to gradient-level injection for SGD) does not correspond to removing $z^*$ from the batch. We therefore inject the perturbation at the gradient level at step $t^*$:
\[
  g_{t^*}(\epsilon) = g_{t^*} - \epsilon\, g_{t^*\!,z^*},
  \qquad
  g_{t^*\!,z^*} := \nabla_\theta \ell(\theta_{t^*}, z^*),
\]
so that $\epsilon = 0$ recovers the original trajectory and $\epsilon = 1$ removes $z^*$. Optimizer states before $t^*$ are unaffected; all subsequent states are functions of $\epsilon$. We use $\dot{(\cdot)}$ to denote the derivative with respect to $\epsilon$ evaluated at $\epsilon = 0$.

\textbf{Auxiliary diagonal matrices.} Throughout we use
\[
  D_t = \operatorname{diag}\!\left(\tfrac{1}{\sqrt{\hat v_t}+\varepsilon}\right),
  \qquad
  S_t = \operatorname{diag}\!\left(\tfrac{\hat m_t}{2\sqrt{\hat v_t}\,(\sqrt{\hat v_t}+\varepsilon)^2}\right),
\]
which arise from differentiating $U_t$ with respect to its inputs.

\subsection{Initial derivatives at step $t^*$}\label{app:adamw-init}

Since all trajectories are identical before step $t^*$, the initial conditions are $\dot\theta_{t^*} = 0$, $\dot m_{t^*-1} = 0$, $\dot v_{t^*-1} = 0$. From the perturbed gradient, $\dot g_{t^*} = -g_{t^*\!,z^*}$.

The moment derivatives at the injection step are:
\begin{align}
  \dot m_{t^*} &= \beta_1 \dot m_{t^*-1} + (1-\beta_1)\dot g_{t^*} = -(1-\beta_1)\,g_{t^*\!,z^*}, \label{eq:app-mdot}\\
  \dot v_{t^*} &= \beta_2 \dot v_{t^*-1} + 2(1-\beta_2)\operatorname{diag}(g_{t^*})\dot g_{t^*} = -2(1-\beta_2)\operatorname{diag}(g_{t^*})\,g_{t^*\!,z^*}. \label{eq:app-vdot}
\end{align}
The bias-corrected derivatives are $\dot{\hat m}_{t^*} = \dot m_{t^*}/(1-\beta_1^{t^*+1})$ and $\dot{\hat v}_{t^*} = \dot v_{t^*}/(1-\beta_2^{t^*+1})$.

Differentiating the update $U_t$ coordinate-wise:
\[
  \dot U_{t^*,i}
  = \eta_{t^*}\!\left(
    \frac{\dot{\hat m}_{t^*,i}}{\sqrt{\hat v_{t^*,i}}+\varepsilon}
    - \frac{\hat m_{t^*,i}\,\dot{\hat v}_{t^*,i}}{2\sqrt{\hat v_{t^*,i}}\,(\sqrt{\hat v_{t^*,i}}+\varepsilon)^2}
  \right).
\]
In vector form,
\[
  \dot U_{t^*} = \eta_{t^*}(D_{t^*}\,\dot{\hat m}_{t^*} - S_{t^*}\,\dot{\hat v}_{t^*}).
\]
Substituting \eqref{eq:app-mdot}--\eqref{eq:app-vdot} and using $\dot\theta_{t^*+1} = -\dot U_{t^*}$:
\begin{equation}\label{eq:app-theta-init}
  \dot\theta_{t^*+1}
  = \eta_{t^*}\!\left(
    \frac{D_{t^*}}{1-\beta_1^{t^*+1}}
    - \frac{2\,S_{t^*}\operatorname{diag}(g_{t^*})}{1-\beta_2^{t^*+1}}
  \right) g_{t^*\!,z^*}.
\end{equation}
The first term arises from the perturbation flowing through the first moment, while the second arises from the second moment. Assembling the three derivatives gives the initial state perturbation:
\begin{equation}\label{eq:app-Zpush}
  Z_{\mathrm{push}}(z^*) = \bigl(\dot\theta_{t^*+1}^\top,\; \dot m_{t^*}^\top,\; \dot v_{t^*}^\top\bigr)^\top \in \mathbb{R}^{3p}.
\end{equation}

\subsection{Recurrence for $t \geq t^*\!+1$}\label{app:adamw-recurrence}

For all subsequent steps, the batch gradient has no direct $\epsilon$-dependence (the only direct perturbation was at $t^*$), so $\dot g_t = \nabla_\theta^2 \ell(\theta_t, \mathcal{B}_t)\,\dot\theta_t = H_t\,\dot\theta_t$. The moment recurrences become
\begin{align*}
  \dot m_t &= \beta_1 \dot m_{t-1} + (1-\beta_1)\,H_t\,\dot\theta_t, \\
  \dot v_t &= \beta_2 \dot v_{t-1} + 2(1-\beta_2)\operatorname{diag}(g_t)\,H_t\,\dot\theta_t,
\end{align*}
and the parameter recurrence is
\[
  \dot\theta_{t+1}
  = (1-\eta_t\lambda)\,\dot\theta_t
    - \eta_t\!\left(
      \frac{D_t}{1-\beta_1^{t+1}}\,\dot m_t
      - \frac{S_t}{1-\beta_2^{t+1}}\,\dot v_t
    \right).
\]

\textbf{Augmented state and linear dynamical system.} Because the gradient-level perturbation propagates jointly through the parameters and both moments, we lift the state to the augmented vector
\[
  Z_t = (\dot\theta_t^\top,\, \dot m_{t-1}^\top,\, \dot v_{t-1}^\top)^\top \in \mathbb{R}^{3p},
\]
on which the perturbation evolves as a linear dynamical system $Z_{t+1} = A_t Z_t$. Collecting the recurrences above, the transition is
\[
  A_t =
  \begin{pmatrix}
    (1\!-\!\eta_t\lambda)I - \frac{\eta_t(1-\beta_1)}{1-\beta_1^{t+1}}D_t H_t + \frac{2\eta_t(1-\beta_2)}{1-\beta_2^{t+1}}S_t\operatorname{diag}(g_t)H_t
    & -\frac{\eta_t\beta_1}{1-\beta_1^{t+1}}D_t
    & \frac{\eta_t\beta_2}{1-\beta_2^{t+1}}S_t \\[6pt]
    (1-\beta_1)H_t & \beta_1 I & 0 \\[4pt]
    2(1-\beta_2)\operatorname{diag}(g_t)H_t & 0 & \beta_2 I
  \end{pmatrix}\!.
\]

\textbf{Decomposition into Hessian-free and Hessian-mediated parts.} The transition admits a clean decomposition $A_t = M_t + R_t H_t P$, where $P = [I_p\;\; 0\;\; 0] \in \mathbb{R}^{p\times 3p}$ extracts the parameter component, and
\begin{subequations}\label{eq:app-MtRt}
\begin{equation}\label{eq:app-Mt}
  M_t =
  \begin{pmatrix}
    (1-\eta_t\lambda)\,I & -\dfrac{\eta_t\beta_1}{1-\beta_1^{t+1}}\,D_t & \dfrac{\eta_t\beta_2}{1-\beta_2^{t+1}}\,S_t \\[6pt]
    0 & \beta_1 I & 0 \\[4pt]
    0 & 0 & \beta_2 I
  \end{pmatrix},
\end{equation}
\begin{equation}\label{eq:app-Rt}
  R_t =
  \begin{pmatrix}
    -\dfrac{\eta_t(1-\beta_1)}{1-\beta_1^{t+1}}\,D_t + \dfrac{2\eta_t(1-\beta_2)}{1-\beta_2^{t+1}}\,S_t\operatorname{diag}(g_t) \\[6pt]
    (1-\beta_1)\,I \\[4pt]
    2(1-\beta_2)\operatorname{diag}(g_t)
  \end{pmatrix}.
\end{equation}
\end{subequations}
$M_t$ collects the Hessian-free terms (momentum decay and weight decay), while $R_t$ collects the Hessian-mediated coupling between optimizer states.

\subsection{Backward recurrence for the summary matrix}\label{app:adamw-backward}

Analogous to SGD-influence, we define a summary matrix $W^{(t)} \in \mathbb{R}^{p\times 3p}$ that encodes how a perturbation injected at step $t$ propagates through all subsequent training steps to the final parameters: $W^{(t)} = P\,\Psi_t$, where $\Psi_t = \prod_{k=t}^{T-1} A_k$.

From $\Psi_t = A_t \Psi_{t+1}$ and the decomposition $A_t = M_t + R_t H_t P$, we obtain
\[
  W^{(t)} = P\,A_t\,\Psi_{t+1} = P(M_t + R_t H_t P)\Psi_{t+1}.
\]
Using the block structure $P M_t = M_t^{(\theta)}$ where $M_t^{(\theta)}$ denotes the first block-row of $M_t$, together with $P\Psi_{t+1} = W^{(t+1)}$, the recurrence becomes
\[
  W^{(t)} = W^{(t+1)} M_t + (W^{(t+1)} R_t)\, H_t\, P,
\]
where we used $P R_t = R_t^{(\theta)}$ and the fact that $W^{(t+1)}$ already absorbs $P$ on the left. Substituting the GGN approximation $H_t \approx \frac{1}{|\mathcal{B}_t|}\sum_{z\in \mathcal{B}_t} g_{t,z}\,g_{t,z}^\top$ yields the per-sample form:
\begin{equation}\label{eq:app-W-update}
  W^{(t)} = W^{(t+1)} M_t + \frac{1}{|\mathcal{B}_t|}\sum_{z\in \mathcal{B}_t}\bigl(W^{(t+1)} R_t\, g_{t,z}\bigr)\,g_{t,z}^\top P,
\end{equation}
with terminal condition $W^{(T)} = [I\;\;0\;\;0]$.

\textbf{Per-sample influence.} The estimated parameter change for sample $z^*$ injected at step $t^*$ is
\[
  \mathrm{AdamW\text{-}influence}_{z^*} = W^{(t^*+1)}\, Z_{\mathrm{push}}(z^*),
\]
combining the initial perturbation \eqref{eq:app-Zpush} with the propagator \eqref{eq:app-W-update}. This recovers the formal version of Proposition~\ref{prop:adamw-influence}: the recurrence \eqref{eq:app-W-update} preserves the $\mathcal{O}(T)$ complexity of SGD-influence, with constant overhead reflecting the augmented state (cost analysis in Appendix~\ref{app:efficient-computation}).

\subsection{Sanity checks}\label{app:adamw-sanity}
 
\paragraph{SGD ($\beta_1=\beta_2=0$, $\lambda=0$).}
Setting $\beta_1=\beta_2=0$ and removing the second moment (i.e., $S_t=0$, $D_t=I$), the transition reduces to $\dot\theta_{t+1} = (I - \eta_t H_t)\dot\theta_t$, recovering the known SGD formula~\citep{hara2019data}.
 
\paragraph{SGD with momentum ($\beta_1>0$, $\beta_2=0$, $\lambda=0$).}
With $\beta_2=0$, $D_t=I$, $S_t=0$, and neglecting bias correction, the system becomes:
\[
  \begin{pmatrix} \dot\theta_{t+1} \\ \dot m_t \end{pmatrix}
  =
  \begin{pmatrix}
    I - \eta_t(1-\beta_1)H_t & -\eta_t\beta_1 I \\
    (1-\beta_1)H_t & \beta_1 I
  \end{pmatrix}
  \begin{pmatrix} \dot\theta_t \\ \dot m_{t-1} \end{pmatrix}\!,
\]
recovering the known SGD-momentum formula~\citep{zhang2025gets}.
 
\subsection{AdamW-influence: algorithm}\label{app:adamw-algorithm}
 
Algorithm~\ref{alg:adamw-dvemb} summarizes the backward recurrent computation of AdamW-influence.
 
\begin{algorithm}[h]
\caption{Backward Recurrent algorithm of AdamW-influence}
\label{alg:adamw-dvemb}
\begin{algorithmic}[1]
\Require Training batches $\{\mathcal{B}_t\}_{t=0}^{T-1}$ with per-sample gradients $\{g_{t,z}\}_{z\in \mathcal{B}_t}$, optimizer states $\{m_t, v_t\}_{t=0}^{T-1}$, learning rates $\{\eta_t\}$, exponential decay rates $\beta_1, \beta_2$, weight decay coefficient $\lambda$, and denominator smoothing term $\varepsilon$.
\Ensure AdamW-influence embeddings $\{\text{AdamW-influence}_z\}_{z \in \mathcal{D}}$
\State $W \leftarrow [I \;\; 0 \;\; 0] \in \mathbb{R}^{p \times 3p}$ \Comment{Initialize summary matrix}
\For{$t = T-1$ \textbf{down to} $0$}
  \State Compute $D_t = \operatorname{diag}(1/(\sqrt{\hat v_t}+\varepsilon))$, \; $S_t = \operatorname{diag}(\hat m_t/(2\sqrt{\hat v_t}(\sqrt{\hat v_t}+\varepsilon)^2))$
  \State Construct $M_t$ from Eq.~\eqref{eq:app-Mt} and $R_t$ from Eq.~\eqref{eq:app-Rt}
  \For{$z \in \mathcal{B}_t$}
    \State $\dot m_t \leftarrow -(1-\beta_1)\,g_{t,z}$ \Comment{Use $g_{t,z}/|\mathcal{B}_t|$ for mean reduction}
    \State $\dot v_t \leftarrow -2(1-\beta_2)\,g_t \odot g_{t,z}$
    \State $\dot{\hat m}_t \leftarrow \dot m_t / (1-\beta_1^{t+1})$, \; $\dot{\hat v}_t \leftarrow \dot v_t / (1-\beta_2^{t+1})$
    \State $\dot\theta_{t+1} \leftarrow -\eta_t(D_t\,\dot{\hat m}_t - S_t\,\dot{\hat v}_t)$
    \State $Z_{\mathrm{push}}(z) \leftarrow (\dot\theta_{t+1}^\top,\, \dot m_t^\top,\, \dot v_t^\top)^\top$
    \State $\text{AdamW-influence}_z \leftarrow W\, Z_{\mathrm{push}}(z)$
  \EndFor
  \State $\Delta W \leftarrow 0$
  \For{$z \in \mathcal{B}_t$}
    \State $v \leftarrow W\, R_t\, g_{t,z}$, \;\; $u^\top \leftarrow (g_{t,z}^\top,\, 0,\, 0)$
    \State $\Delta W \leftarrow \Delta W + v\, u^\top$
  \EndFor
  \State $W \leftarrow W\, M_t + \frac{1}{|\mathcal{B}_t|}\,\Delta W$
\EndFor
\State \Return $\{\text{AdamW-influence}_z\}_{z \in \mathcal{D}}$
\end{algorithmic}
\end{algorithm}

\section{Efficient computation and storage of AdamW-influence}\label{app:efficient-computation}

\subsection{Random masking and ensembles}
\label{sec:random_masking}

The per-sample gradient and momentums cached by AdamW-influence live in $\mathbb{R}^p$, where $p$ is the number of model parameters. Storing and manipulating $p$-dimensional embedding per training sample is prohibitive at modern scale, so dimensionality reduction is essential.

SGD-influence~\citep{wang2025capturing} admits an efficient random-projection implementation because its backward recurrence is linear in the per-sample gradients, so the Johnson--Lindenstrauss lemma preserves the relevant inner products under a random projection. AdamW-influence breaks this property: the matrices $D_t$ and $S_t$ in Eqs.~\ref{eq:app-MtRt} apply a coordinate-wise nonlinearity to $\hat{v}_t$, which a random projection scrambles.

\paragraph{Random masking.} We sample once a fixed binary mask $\mathbf{m} \in \{0,1\}^p$, and restrict the entire AdamW-influence computation to the index set $\mathcal{S} = \{i : \mathbf{m}_i = 1\}$. All per-sample gradients $g_{t,z}$, batch gradients $g_t$, and optimizer states are subselected to $\mathcal{S}$ so Algorithm~\ref{alg:adamw-dvemb} remains the same with $p$ replaced by $|\mathcal{S}|$. The mask is shared across all training steps and samples within a single run.

\paragraph{Ensembles for masking.} At the LLM scale, memory constraints push $|\mathcal{S}| / p$ to small values, where a single mask produces significant variance and information loss. We mitigate this with an ensemble of $M$ independent masks, run AdamW-influence once per mask to obtain scores, and aggregate by simple average. Per-setting choices of $M$ are reported in Appendix~\ref{app:exp-details}.

\subsection{Computational efficiency}
\label{sec:computational_efficiency}

We compare the cost of AdamW-influence against SGD-influence along three axes: per-sample embedding storage, working memory during the backward recurrence, and per-step compute. Throughout this section, $p$ denotes the number of model parameters, $T$ the number of training steps, $b$ the batch size, and $|\mathcal{S}|$ the number of coordinates retained after random masking ($|\mathcal{S}| = p$ if masking is disabled). For ease of exposition, we first describe costs in terms of $p$ and then note that masking replaces $p$ with $|\mathcal{S}|$ throughout, providing the same multiplicative cost reduction as random projection in SGD-influence.

\paragraph{Working memory and storage.} The summary matrix $W^{(t)} \in \mathbb{R}^{p \times 3p}$ is $3\times$ larger than the $\mathbb{R}^{p \times p}$ matrix in SGD-influence, reflecting the extended state $(\dot{\theta}, \dot{m}, \dot{v})$ that AdamW-influence must propagate. Per-step trajectory caching incurs a similar additive overhead: AdamW-influence stores $\hat{m}_t, \hat{v}_t$ alongside the per-sample gradients $\{g_{t,z}\}_{z \in \mathcal{B}_t}$ and batch gradient $g_t$, adding $2p$ per step on top of the $bp + p$ already required by SGD-influence. When $b$ is not too small, this overhead is dominated by the $bp$ per-sample gradient term and is $\mathcal{O}(1/b)$ relative to it.

\paragraph{Per-step compute.} The dominant cost in the backward recurrence is the update of $W^{(t)}$ in Eq.~\ref{eq:app-W-update}. The term $W^{(t+1)} M_t$ exploits the block structure of $M_t$ (each block is either diagonal or a scalar multiple of identity), so it costs $\mathcal{O}(p^2)$. The per-sample term $W^{(t+1)} R_t g_{t,z}$ similarly leverages the block structure of $R_t$, costing $\mathcal{O}(p^2)$ per sample, and the rank-one update to $W^{(t)}$ costs $\mathcal{O}(3p^2)$ per sample due to the wider second dimension. The total per-step cost is $\mathcal{O}(3bp^2)$, compared to $\mathcal{O}(bp^2)$ for SGD-influence, a $3\times$ overhead arising directly from the wider summary matrix.

\paragraph{Effect of random masking.} With masking, every $p$ above is replaced by $|\mathcal{S}|$. The $3\times$ ratios between AdamW-influence and SGD-influence are preserved. Random masking thus delivers the same form of cost reduction that random projection provides in SGD-influence---turning otherwise infeasible quadratic costs in $p$ into tractable quadratic costs in $|\mathcal{S}|$---without breaking the coordinate-wise structure required by AdamW (Section~\ref{sec:random_masking}). In summary, AdamW-influence incurs a $3\times$ overhead in working memory and per-step compute relative to SGD-influence, while keeping the per-sample embedding size identical.

\begin{table}[h]
\centering
\caption{Cost comparison between SGD-influence and AdamW-influence. The first four rows describe the per-component costs without masking; the last two row gives the total cost when random masking without/with $|\mathcal{S}|$ retained coordinates is applied. The $3\times$ ratios are preserved under masking.}
\label{tab:cost_comparison}
\begin{tabular}{lccc}
\toprule
& SGD-influence & AdamW-influence & Ratio (Adam / SGD) \\
\midrule
Per-sample embedding (storage) & $p$ & $p$ & $1\times$ \\
Summary matrix $W$ (memory) & $p^2$ & $3p^2$ & $3\times$ \\
Trajectory cache (per step) & $bp + p$ & $bp + 3p$ & $\approx 1\times$ \\
Backward recurrence (per step) & $\mathcal{O}(bp^2)$ & $\mathcal{O}(3bp^2)$ & $3\times$ \\
\midrule
\textbf{Total cost} & $\mathcal{O}(Tbp^2)$ & $\mathcal{O}(3Tbp^2)$ & $3\times$ \\
\textbf{Total cost (with masking $|\mathcal{S}|$)} & $\mathcal{O}(Tb|\mathcal{S}|^2)$ & $\mathcal{O}(3Tb|\mathcal{S}|^2)$ & $3\times$ \\
\bottomrule
\end{tabular}
\end{table}

\section{Experiment details}\label{app:exp-details}

\subsection{Experimental settings for fidelity evaluation}~\label{app:fidelity-settings}

In section~\ref{sec:fidelity}, we consider four experiment settings to evaluate the attribution fidelity of AdamW-influence. We also consider additional settings presented in Appendix~\ref{app:additional-exp-results}.

\begin{itemize}
    \item \textbf{MNIST+MLP.} We train a three-layer multilayer perceptron (MLP) with hidden layer sizes of 16, consisting of two fully connected layers ($28\times28 \rightarrow 16 \rightarrow 16$) with ReLU activations, followed by a final linear layer mapping to 10 output classes on the MNIST dataset~\citep{lecun2002gradient}. We train on 10\% of MNIST for 1 epoch. We set the batch size to 64 and perform the experiment on 5 learning rates ($10^{-2}$, $10^{-3}$, $10^{-4}$, $10^{-5}$ and $10^{-6}$). TSLOO ground truth is computed for 200 randomly selected training samples. The Spearman correlation is computed across these 200 samples and averaged over 500 validation points. We use AdamW, we set $\beta_1 = 0.9$ and $\beta_2 = 0.95$.
    
    \item \textbf{MNIST+CNN.} We train a convolutional neural network (CNN) with two convolutional layers (Conv2d(1, 32, 3, padding=1) and Conv2d (32, 64, 3, padding=1), each followed by ReLU and 2×2 max pooling) and a final fully connected layer of size $64\times7\times7$ $\rightarrow$ 10 on the MNIST dataset~\citep{lecun2002gradient}. We train on 10\% of MNIST for one epoch. We set the batch size to 64 and perform the experiment on 5 learning rates ($10^{-2}$, $10^{-3}$, $10^{-4}$, $10^{-5}$, $10^{-6}$). We use a random masking ratio of 0.75 with a single masking ensemble. TSLOO ground truth is computed for 200 randomly selected training samples. The Spearman correlation is computed across these 200 samples and averaged over 500 validation points. We use AdamW, we set $\beta_1 = 0.9$ and $\beta_2 = 0.95$.
    
    \item \textbf{WikiText-2+GPT-2.} We continually pretrain a GPT-2 language model \cite{radford2019language} on the WikiText-2 dataset \cite{merity2016pointer} by splitting the corpus into non-overlapping token blocks. GPT-2 is licensed under MIT license and WikiText-2 is licensed under CC BY-SA 3.0. We experiment with multiple training configurations, changing the number of training samples and block sizes, specifically: 512 and 1024 training samples with block sizes of 128 and 256. All models are trained for three epochs. We use the AdamW optimizer with three learning rates ($1 \times10^{-4}$, $5 \times 10^{-5}$, and $1 \times10^{-5}$) and a linear learning rate scheduler with warmup. We set the batch size to 32 and use a random masking ensemble of size 10 with projection dimension $512$. TSLOO ground truth is computed for 50 randomly selected training samples. The Spearman correlation is computed across these 50 samples and averaged over 256 validation points.

    \item \textbf{Alpaca + Llama 3.2-1B.} We fine-tune a LLaMA-3.2-1B language model~\citep{grattafiori2024llama} on the Alpaca instruction-tuning dataset~\citep{alpaca}, using the first $512$ training examples with a maximum sequence length of $512$ tokens. We train for $1$ epoch and evaluate attribution fidelity on $100$ test samples from the SAMSum summarization task~\citep{gliwa-etal-2019-samsum}. We use the AdamW optimizer and sweep learning rates spanning $5\times10^{-7}$, $2\times10^{-6}$, and $6\times10^{-6}$  ($\beta_1=0.9$, $\beta_2=0.999$, max gradient norm 1.0) and a linear learning-rate scheduler with $100$ warmup steps. We evaluate both AdamW-influence and SGD-influence. We set the batch size to $8$ and use a random-masking ensemble of size $10$ with projection dimension $256$. TSLOO ground truth is computed for 50 randomly selected training samples. The Spearman correlation is computed across these 50 samples; attribution scores are then compared against TSLOO effects on 256 SAMSum~\citep{gliwa-etal-2019-samsum} test samples.
    
\end{itemize}

\subsection{Experimental settings for dissecting algorithm-level error}

In Section~\ref{sec:approx-error}, we conduct three analyses on the algorithm-level error of AdamW-influence: (i) decomposing the absolute SGD-influence error into config-level, algorithm-level, and residual components (Section~\ref{ssec:error-accumulation}, Figure~\ref{fig:error-decomposition}); (ii) sweeping learning rates to identify the two governing factors of the algorithm-level error (Section~\ref{sec:two-factors}, Figure~\ref{fig:two-factors}); and (iii) validating the closed-form error proxy against ground-truth update-estimation error norms (Section~\ref{sec:error-proxy}, Figure~\ref{fig:proxy-validation}). All experiments are conducted on the MNIST+MLP and MNIST+CNN settings; model architecture, dataset preparation, batch size, optimizer hyperparameters ($\beta_1=0.9$, $\beta_2=0.95$), random masking configuration, and the number of LOO training samples (200) and TSLOO validation samples (500) follow Appendix~\ref{app:fidelity-settings}. Below we describe only the analysis-specific details.

\begin{itemize}
    \item \textbf{Error decomposition (Section~\ref{ssec:error-accumulation}, Figure~\ref{fig:error-decomposition}).} We use the MNIST+MLP setting and sweep three learning rates ($10^{-3}$, $10^{-4}$, $10^{-5}$). For each sample $z^*$, we compute its absolute SGD-influence error against the TSLOO ground truth and decompose it into the three additive components in Equation~\eqref{eq:proxy-unroll}: the optimizer-mismatch share (green), the update-estimation error remaining after optimizer alignment (blue), and the higher-order residual (grey). Errors from samples whose injection step $t^*$ falls within the same 5-step window are aggregated into a single bin, yielding the binned curves shown in Figure~\ref{fig:error-decomposition}.
    
    \item \textbf{Sweep of governing factors (Section~\ref{sec:two-factors}, Figure~\ref{fig:two-factors}).} We use the MNIST+MLP and MNIST+CNN settings and sweep three learning rates ($10^{-3}$, $10^{-4}$, $10^{-5}$). For each $(z^*, t^*)$ pair, we compute two metrics: (i) the parameter-space error norm $\|\Delta\theta_{z^*} - \text{AdamW-influence}_{z^*}\|_2$, where $\Delta\theta_{z^*}$ is obtained by trajectory-specific LOO retraining; and (ii) the intra-step Spearman correlation between AdamW-influence scores and ground-truth TSLOO scores computed over all samples sharing the same injection step $t^*$. The intra-step correlation curves in Figure~\ref{fig:two-factors}(b, d) are smoothed by a rolling mean over a window of 5 consecutive training steps to reduce step-to-step noise; the error-norm curves in Figure~\ref{fig:two-factors}(a, c) are reported without smoothing.
    
    \item \textbf{Error proxy validation (Section~\ref{sec:error-proxy}, Figure~\ref{fig:proxy-validation}).} We use the MNIST+MLP and MNIST+CNN settings, each at two learning rates ($10^{-3}$ and $10^{-4}$). For every training sample (4992 in total per setting), we compute the closed-form error proxy from Equation~\eqref{eq:proxy-unroll} along the original training trajectory, alongside the ground-truth update-estimation error norm $\|\Delta\theta_{z^*} - \text{AdamW-influence}_{z^*}\|_2$ obtained via trajectory-specific LOO retraining. Each point in Figure~\ref{fig:proxy-validation} corresponds to one training sample; the reported Spearman $\rho$, slope, and $R^2$ are computed across all 4992 samples per panel.
\end{itemize}

\subsection{Experimental settings for data selection}
\label{app:data-selection-settings}

In Section~\ref{sec:practical-data-sel}, we evaluate the unified $K$-step look-ahead framework on four settings: MNIST+MLP, MNIST+CNN, Tulu3+Llama 3.2-1B, and Alpaca+Llama 3.2-1B. Across all three settings, we compare three methods: \emph{offline AdamW-influence}, \emph{online SGD-influence}, and \emph{online AdamW-influence}. Offline AdamW-influence does not involve $K$, since it scores candidates against a completed reference trajectory and retrains from scratch on the selected subset. For online methods, the look-ahead horizon $K$ is swept and the best $K$ per setting (selected by validation performance) is reported in Figure~\ref{fig:offline-adam-sgd-datasel}: $K=10$ for MNIST+MLP, $K=2$ for Tulu3+Llama 3.2-1B, and $K=2$ for Alpaca+Llama 3.2-1B. Section~\ref{sec:practical implications} additionally sweeps $K \in \{2, 5, 10, 25\}$ across three learning rates on MNIST+MLP (Figure~\ref{fig:selection-k-sweep}). Below we describe the per-setting details.

\begin{itemize}
    \item \textbf{MNIST+MLP.} Model architecture and dataset preparation follow Appendix~\ref{app:fidelity-settings}. We train for 20 epochs (rather than 1 epoch as in fidelity evaluation) to allow the model to converge under the reduced effective batch size induced by selection. At each training step, the candidate batch size is $N = 64$ and the final batch size after selection is $B = 32$, corresponding to a selected ratio of $0.5$. For Figure~\ref{fig:offline-adam-sgd-datasel}, we use learning rate $10^{-3}$ and $K = 10$. For the $K$-sweep in Figure~\ref{fig:selection-k-sweep}, we sweep $K \in \{2, 5, 10, 25\}$ across three learning rates $\{10^{-2}, 10^{-3}, 10^{-4}\}$. Validation and test error rates are reported at the best validation epoch.

    \item \textbf{MNIST+CNN.} Model architecture and dataset preparation follow Appendix~\ref{app:fidelity-settings}. We train for 20 epochs (rather than 1 epoch as in fidelity evaluation) to allow the model to converge under the reduced effective batch size induced by selection. At each training step, the candidate batch size is $N = 64$ and the final batch size after selection is $B = 32$, corresponding to a selected ratio of $0.5$. For Figure~\ref{fig:offline-adam-sgd-datasel}, we use learning rate $10^{-3}$ and $K = 5$. For the $K$-sweep in Figure~\ref{fig:selection-k-sweep}, we sweep $K \in \{2, 5, 10\}$ across three learning rates $\{10^{-2}, 10^{-3}, 10^{-4}\}$. Validation and test error rates are reported at the best validation epoch.

    \item \textbf{Alpaca+Llama 3.2-1B.} Model, optimizer hyperparameters ($\beta_1=0.9$, $\beta_2=0.999$, max gradient norm $1.0$), maximum sequence length (512 tokens), training batch size (8), random-masking ensemble of size 10 with projection dimension 256, and linear learning-rate scheduler with 100 warmup steps follow Appendix~\ref{app:fidelity-settings}. We deviate from the fidelity setup as follows: (i) we use 40\% of the Alpaca dataset for training (rather than the first 512 examples); (ii) we use a learning rate of $2 \times 10^{-6}$; (iii) full fine-tuning is used (no LoRA); (iv) the candidate batch size is $N = 8$ and the selected ratio is $0.8$, so $B = 6$ samples are retained per step; and (v) validation uses 16 randomly sampled SAMSum examples for online method selection, and final evaluation is reported on the full SAMSum test set. Validation and test perplexity are reported at the final training step.

    \item \textbf{Tulu3+Llama 3.2-1B.} The training setup mirrors the Alpaca+Llama 3.2-1B configuration above, with the following differences: (i) we use 1\% of the Tulu3 dataset for training; (ii) the learning rate is $4 \times 10^{-5}$ for full fine-tuning (used in Figure~\ref{fig:offline-adam-sgd-datasel}) and $3 \times 10^{-3}$ for LoRA fine-tuning (used only in Tables~\ref{tab:tulu3_tydiqa_eval} and~\ref{tab:tulu3_tydiqa_val} of Appendix~\ref{app:additional-data-sel}); (iii) the candidate batch size is $N = 8$ and the selected ratio is $0.5$, so $B = 4$ samples are retained per step; and (iv) validation uses 16 randomly sampled tydiqa~\citep{clark2020tydi} examples for online method selection, and final evaluation is reported on the full tydiqa test set. Validation and test perplexity are reported at the final training step.

    \item \textbf{Additional baseline in Tables~\ref{tab:tulu3_tydiqa_eval} and~\ref{tab:tulu3_tydiqa_val}.} The Tulu3+Llama 3.2-1B results in Appendix~\ref{app:additional-data-sel} additionally include TracIn~\citep{pruthi2020estimating} as an offline baseline, and sweep both fine-tuning regime (full vs.\ LoRA) and selected ratio ($0.5$ vs.\ $0.8$). All other training and evaluation details follow the Tulu3+Llama 3.2-1B configuration above.

\end{itemize}

\subsection{Compute Resources}\label{app:compute-resources}

The experiment is carried out on A100 GPUs. The total GPU hour required for the experiments is around 400 hours.

\section{Additional Experiment Results}\label{app:additional-exp-results}

\subsection{Fidelity additional results}\label{app:additional-fidelity}

Tables~\ref{tab:fidelity-full} and~\ref{tab:fidelity-sgd-training} together demonstrate that attribution fidelity hinges on aligning the attribution formula with the actual training optimizer. Table~\ref{tab:fidelity-full} extends Table~\ref{tab:fidelity-adamw} to additional data sizes and learning rates: when training with AdamW, AdamW-influence consistently outperforms SGD-influence across all four settings, with relative improvements ranging from 10\% to over 300\%. The improvement is especially large at higher learning rates, where the SGD update fails most dramatically as a proxy for the AdamW update. We also reported the error bar under 10 independent runs for MNIST experiments and 5 independe runs for GPT-2 / Llama 3.2 experiments.

Table~\ref{tab:fidelity-sgd-training} reports the symmetric sanity check: when the model is instead trained with SGD, SGD-influence dominates AdamW-influence by a similarly wide margin. This is the expected behavior. Under SGD training, the cached optimizer states $g_t, m_t, v_t$ that AdamW-influence relies on are computed along an SGD trajectory rather than an AdamW one, so AdamW-influence is unrolling an optimizer that the model never actually used. The result confirms that the gains in Table~\ref{tab:fidelity-full} come from \emph{matching} the attribution formula to the training optimizer, not from any intrinsic superiority of AdamW-influence over SGD-influence: each method attributes faithfully only when its assumed optimizer matches the one used in training.

\begin{table}[h]
\centering
\caption{Full fidelity comparison (Spearman $\rho$ against TSLOO ground truth) between AdamW-influence and SGD-influence when training with \textbf{AdamW}, across varying epochs, data sizes, and learning rates. Best result per row in \textbf{bold}. $\Delta$ is the relative improvement of AdamW-influence over SGD-influence.}\label{tab:fidelity-full}
\resizebox{\textwidth}{!}{%
\begin{tabular}{lllcccc}
\toprule
Setting & Epochs & Data Size & LR & AdamW-influence & SGD-influence & $\Delta$ \\
\midrule

\multirow{5}{*}{MNIST+MLP}
& \multirow{5}{*}{1} 
& \multirow{5}{*}{4992} 
& 1e-2 & $\mathbf{0.043{\pm}0.003}$ & $0.023{\pm}0.004$ & $+87\%$ \\
& & & 1e-3 & $\mathbf{0.205{\pm}0.011}$ & $0.075{\pm}0.009$ & $+173\%$ \\
& & & 1e-4 & $\mathbf{0.294{\pm}0.014}$ & $0.242{\pm}0.016$ & $+21\%$ \\
& & & 1e-5 & $\mathbf{0.786{\pm}0.013}$ & $0.715{\pm}0.012$ & $+10\%$ \\
& & & 1e-6 & $\mathbf{0.948{\pm}0.006}$ & $0.833{\pm}0.011$ & $+14\%$ \\
\cmidrule(lr){1-7}

\multirow{5}{*}{MNIST+CNN}
& \multirow{5}{*}{1} 
& \multirow{5}{*}{4992} 
& 1e-2 & $\mathbf{0.048{\pm}0.008}$ & $0.027{\pm}0.011$ & $+74\%$ \\
& & & 1e-3 & $\mathbf{0.090{\pm}0.008}$ & $0.015{\pm}0.009$ & $+500\%$ \\
& & & 1e-4 & $\mathbf{0.511{\pm}0.017}$ & $0.122{\pm}0.015$ & $+319\%$ \\
& & & 1e-5 & $\mathbf{0.791{\pm}0.007}$ & $0.526{\pm}0.007$ & $+50\%$ \\
& & & 1e-6 & $\mathbf{0.767{\pm}0.013}$ & $0.523{\pm}0.009$ & $+47\%$ \\
\cmidrule(lr){1-7}

\multirow{9}{*}{WikiText+GPT-2}
& \multirow{9}{*}{3}

& \multirow{3}{*}{65K (block size= 128)}
& 1e-4 & $\mathbf{0.734{\pm}0.003}$ & $0.372{\pm}0.006$ & $+97\%$ \\
& & & 5e-5 & $\mathbf{0.825{\pm}0.002}$ & $0.432{\pm}0.009$ & $+91\%$ \\
& & & 1e-5 & $\mathbf{0.842{\pm}0.006}$ & $0.480{\pm}0.015$ & $+75\%$ \\
\cmidrule(lr){3-7}

& & \multirow{3}{*}{131K (block size= 256)}
& 1e-4 & $\mathbf{0.672{\pm}0.009}$ & $0.321{\pm}0.011$ & $+109\%$ \\
& & & 5e-5 & $\mathbf{0.787{\pm}0.003}$ & $0.383{\pm}0.011$ & $+105\%$ \\
& & & 1e-5 & $\mathbf{0.810{\pm}0.009}$ & $0.420{\pm}0.011$ & $+93\%$ \\
\cmidrule(lr){3-7}

& & \multirow{3}{*}{262K (block size= 256)}
& 1e-4 & $\mathbf{0.327{\pm}0.013}$ & $0.158{\pm}0.013$ & $+107\%$ \\
& & & 5e-5 & $\mathbf{0.471{\pm}0.012}$ & $0.223{\pm}0.009$ & $+111\%$ \\
& & & 1e-5 & $\mathbf{0.785{\pm}0.008}$ & $0.393{\pm}0.007$ & $+100\%$ \\
\cmidrule(lr){1-7}

\multirow{3}{*}{Alpaca+Llama~3.2}
& \multirow{3}{*}{1} 
& \multirow{3}{*}{512} 
& 6e-6 & $\mathbf{0.276{\pm}0.047}$ & $0.156{\pm}0.019$ & $+77\%$ \\
& & & 2e-6 & $\mathbf{0.342{\pm}0.036}$ & $0.197{\pm}0.007$ & $+74\%$ \\
& & & 5e-7 & $\mathbf{0.665{\pm}0.013}$ & $0.330{\pm}0.017$ & $+102\%$ \\

\bottomrule
\end{tabular}%
}
\end{table}

\begin{table}[ht]
\centering
\caption{Fidelity comparison (Spearman $\rho$ against TSLOO ground truth) between AdamW-influence and SGD-influence when training with \textbf{SGD}, across varying epochs, data sizes, and learning rates. Best result per row in \textbf{bold}.}\label{tab:fidelity-sgd-training}
\resizebox{0.8\textwidth}{!}{%
\begin{tabular}{lllccc}
\toprule
Setting & Epochs & Data Size & LR & AdamW-influence & SGD-influence \\
\midrule
\multirow{5}{*}{MNIST+MLP}
& \multirow{5}{*}{1}
& \multirow{5}{*}{4992}
& 1e-2 & $0.015{\pm}0.004$ & $\mathbf{0.349{\pm}0.016}$ \\
& & & 1e-3 & $0.047{\pm}0.005$ & $\mathbf{0.707{\pm}0.023}$ \\
& & & 1e-4 & $0.447{\pm}0.017$ & $\mathbf{0.939{\pm}0.012}$ \\
& & & 1e-5 & $0.746{\pm}0.015$ & $\mathbf{0.968{\pm}0.001}$ \\
& & & 1e-6 & $0.833{\pm}0.010$ & $\mathbf{0.969{\pm}0.001}$ \\
\cmidrule(lr){1-6}
\multirow{5}{*}{MNIST+CNN}
& \multirow{5}{*}{1}
& \multirow{5}{*}{4992}
& 1e-2 & $0.042{\pm}0.004$ & $\mathbf{0.389{\pm}0.015}$ \\
& & & 1e-3 & $0.036{\pm}0.006$ & $\mathbf{0.808{\pm}0.009}$ \\
& & & 1e-4 & $0.061{\pm}0.006$ & $\mathbf{0.599{\pm}0.011}$ \\
& & & 1e-5 & $0.461{\pm}0.009$ & $\mathbf{0.613{\pm}0.015}$ \\
& & & 1e-6 & $0.480{\pm}0.010$ & $\mathbf{0.616{\pm}0.016}$ \\
\bottomrule
\end{tabular}%
}
\end{table}

\subsection{Data selection additional results}\label{app:additional-data-sel}

Tables~\ref{tab:tulu3_tydiqa_eval} and~\ref{tab:tulu3_tydiqa_val} extend the Tulu3 results in Section~\ref{sec:practical implications} along two practical axes: the fine-tuning regime (full fine-tuning vs.\ LoRA) and the selected data ratio ($0.5$ vs.\ $0.8$). Online AdamW-influence achieves the lowest perplexity across all eight configurations on both validation and test splits.

\begin{table}[h]
\centering
\caption{Evaluation perplexity on Tulu3 (tydiqa) with Llama-3.2-1B. Lower is better; \textbf{bold} indicates the best result per column.}
\label{tab:tulu3_tydiqa_eval}
\begin{tabular}{lcccc}
\toprule
& \multicolumn{2}{c}{Full} & \multicolumn{2}{c}{LoRA} \\
\cmidrule(lr){2-3} \cmidrule(lr){4-5}
Method $\backslash$ Selected Ratio & 0.5 & 0.8 & 0.5 & 0.8 \\
\midrule
TracIN (offline)         & 2.152$\pm$0.063 & 2.175$\pm$0.081 & 1.955$\pm$0.027 & 1.910$\pm$0.065 \\
SGD-influence (online)   & 2.029$\pm$0.034 & 2.055$\pm$0.077 & 1.810$\pm$0.026 & 1.872$\pm$0.016 \\
AdamW-influence (offline) & 1.957$\pm$0.021 & 1.939$\pm$0.017 & 1.868$\pm$0.028 & 1.872$\pm$0.061 \\
AdamW-influence (online)  & \textbf{1.886$\pm$0.045} & \textbf{1.933$\pm$0.018} & \textbf{1.764$\pm$0.021} & \textbf{1.859$\pm$0.012} \\
\bottomrule
\end{tabular}
\end{table}

\begin{table}[h]
\centering
\caption{Validation perplexity on Tulu3 (tydiqa) with Llama-3.2-1B. Lower is better; \textbf{bold} indicates the best result per column.}
\label{tab:tulu3_tydiqa_val}
\begin{tabular}{lcccc}
\toprule
& \multicolumn{2}{c}{Full} & \multicolumn{2}{c}{LoRA} \\
\cmidrule(lr){2-3} \cmidrule(lr){4-5}
Method $\backslash$ Selected Ratio & 0.5 & 0.8 & 0.5 & 0.8 \\
\midrule
TracIN (offline)         & 2.066$\pm$0.087 & 2.154$\pm$0.140 & 1.811$\pm$0.035 & 1.872$\pm$0.069 \\
SGD-influence (online)   & 1.849$\pm$0.057 & 1.857$\pm$0.022 & 1.729$\pm$0.017 & 1.730$\pm$0.054 \\
AdamW-influence (offline) & 1.836$\pm$0.042 & 1.874$\pm$0.018 & 1.803$\pm$0.042 & 1.836$\pm$0.062 \\
AdamW-influence (online)  & \textbf{1.742$\pm$0.025} & \textbf{1.776$\pm$0.041} & \textbf{1.707$\pm$0.042} & \textbf{1.723$\pm$0.063} \\
\bottomrule
\end{tabular}
\end{table}

\section{Derivation of Proposition~\ref{prop:proxy}}\label{app:error-proxy-derivation}

This appendix derives Proposition~\ref{prop:proxy} (the closed-form error proxy in Section~\ref{sec:error-proxy}). All notation follows Appendix~\ref{app:adamw-notation}: $D_t$ and $S_t$ are the diagonal matrices defined there, and $\dot{(\cdot)}$ denotes derivatives with respect to the gradient-level perturbation scalar $\epsilon$ at $\epsilon = 0$. We treat $\lambda = 0$; the extension to $\lambda > 0$ replaces each $(I - A_k^{\mathrm{eff}})$ below with $((1 - \eta_k \lambda) I - A_k^{\mathrm{eff}})$ and leaves $r_t(0)$ unchanged.

\subsection{Error recursion}\label{app:proxy-recursion}

Recall the AdamW update $\theta_{t+1}(\epsilon) = \theta_t(\epsilon) - U_t(\epsilon)$ with $U_t = \eta_t \hat m_t / (\sqrt{\hat v_t} + \varepsilon)$, and its linearization $\dot\theta_{t+1} = \dot\theta_t - \dot U_t$. Define
\begin{equation*}
e_t := [\theta_t(1) - \theta_t(0)] - \dot\theta_t,
\qquad
r_t := [U_t(1) - U_t(0)] - \dot U_t.
\end{equation*}
Subtracting the linearized recursion from the true one gives $e_{t+1} = e_t - r_t$ with $e_{t^*} = 0$.

Although $U_t(1)$ is evaluated on the ground-truth retrained trajectory $\theta_t(1)$, expressing $\theta_t(1) = \theta_t(0) + \dot\theta_t + e_t$ and Taylor-expanding $U_t$ around $\theta_t(0)$ introduces an $e_t$-dependence:
\begin{equation*}
r_t(e_t) \approx r_t(0) + A_t^{\mathrm{eff}}\, e_t,
\qquad
A_t^{\mathrm{eff}} := \left.\frac{\partial U_t}{\partial \theta_t}\right|_{\theta_t(0)}
= \eta_t\, \frac{1 - \beta_1}{1 - \beta_1^{t+1}}\, D_t H_t - 2 \eta_t\, \frac{1 - \beta_2}{1 - \beta_2^{t+1}}\, S_t\, \mathrm{diag}(g_t)\, H_t,
\end{equation*}
where $r_t(0)$ is the residual evaluated on the unperturbed trajectory and $A_t^{\mathrm{eff}}$ follows from differentiating $U_t$ through the moment recurrences. Substituting yields $e_{t+1} = (I - A_t^{\mathrm{eff}})\, e_t - r_t(0)$, which unrolls to
\begin{equation}\label{eq:eT-unrolled}
e_T = -\sum_{t = t^*}^{T-1} \left[\prod_{k = t+1}^{T-1} (I - A_k^{\mathrm{eff}})\right] r_t(0).
\end{equation}
When $\|A_t^{\mathrm{eff}}\| \ll 1$, the propagation factor $\prod_k (I - A_k^{\mathrm{eff}}) \approx I$ and the proxy simplifies to $e_T \approx -\sum_t r_t(0)$, which we use to compute Figure~\ref{fig:proxy-validation}.

\subsection{Magnitude of $r_t(0)$}\label{app:proxy-magnitude}

Writing $U_{t,i} = \eta_t f_{t,i}$ with $f_{t,i}(\epsilon) = \hat m_{t,i}(\epsilon) / (\sqrt{\hat v_{t,i}(\epsilon)} + \varepsilon)$, we have $r_{t,i}(0) = \tfrac{1}{2} \eta_t \ddot f_{t,i}|_{\epsilon = 0}$. Let $h_i := \sqrt{\hat v_{t,i}} + \varepsilon$. Direct differentiation gives
\begin{equation}\label{eq:fddot}
\ddot f_{t,i}
= \frac{\ddot{\hat m}_{t,i}}{h_i}
- \frac{\hat m_{t,i}\, \ddot h_i}{h_i^2}
- \frac{2\, \dot{\hat m}_{t,i}\, \dot h_i}{h_i^2}
+ \frac{2\, \hat m_{t,i}\, \dot h_i^2}{h_i^3},
\end{equation}
with $\dot h_i = \dot{\hat v}_{t,i} / (2 \sqrt{\hat v_{t,i}})$ and $\ddot h_i = \ddot{\hat v}_{t,i} / (2 \sqrt{\hat v_{t,i}}) - \dot{\hat v}_{t,i}^2 / (4 \hat v_{t,i}^{3/2})$.

At $t > t^*$, the moment derivatives admit the leading-order expressions (suppressing bias-correction prefactors):
\begin{equation*}
\dot{\hat m}_{t,i} \sim [H_t \dot\theta_t]_i,
\qquad
\ddot{\hat m}_{t,i} \sim \sum_{j,k} [\nabla^3 \ell_t]_{ijk}\, \dot\theta_{t,j}\, \dot\theta_{t,k} \sim \|\dot\theta_t\|^2,
\qquad
\dot h_i \sim \frac{g_{t,i}\, \dot g_{t,i}}{\sqrt{\hat v_{t,i}}},
\qquad
\ddot h_i \sim \frac{\dot g_{t,i}^2}{\sqrt{\hat v_{t,i}}}.
\end{equation*}
Using $h_i \approx \sqrt{\hat v_{t,i}}$ and the Adam normalization $\hat m_{t,i} / \sqrt{\hat v_{t,i}} = \mathcal{O}(1)$, the four terms in~\eqref{eq:fddot} scale as $\|\dot\theta_t\|^2 / \sqrt{\hat v_{t,i}}$, $[H_t \dot\theta_t]_i^2 / \hat v_{t,i}$, and (for terms 3 and 4) $[H_t \dot\theta_t]_i^2 / \hat v_{t,i}$ further suppressed by $|g_{t,i}| / \sqrt{\hat v_{t,i}} \leq 1$ (the batch-mean gradient is bounded above by the per-coordinate RMS accumulated by Adam). Terms 1 and 2 dominate, giving
\begin{equation}\label{eq:rt-bound}
\bigl|r_{t,i}(0)\bigr|
= \tfrac{1}{2} \eta_t \bigl|\ddot f_{t,i}\bigr|
\sim \eta_t \!\left(
  \frac{\|\dot\theta_t\|^2}{\sqrt{\hat v_{t,i}}}
  + \frac{[H_t \dot\theta_t]_i^2}{\hat v_{t,i}}
\right),
\end{equation}
which recovers the residual bound stated in Proposition~\ref{prop:proxy}. The first term originates from the loss nonlinearity ($\nabla^3 \ell$) entering $\ddot{\hat m}_{t,i}$ and is present even for non-adaptive optimizers; the second originates from the quadratic dependence of $\hat v_t$ on gradient perturbations and is specific to Adam-family optimizers. Both belong to a single second-order remainder of $f_{t,i}$ and are not independent error sources.

\section{Pseudo-code of the unified $K$-step selection loop}
\label{app:kstep-pseudocode}

Algorithm~\ref{alg:kstep} summarizes the unified $K$-step look-ahead data selection loop introduced in Section~\ref{sec:unified-K}.

\begin{algorithm}[h]
\caption{$K$-step look-ahead data selection}
\label{alg:kstep}
\begin{algorithmic}[1]
\Require Candidate batch size $N$; training batch size $B$; look-ahead horizon $K$; validation set $V$
\State Initialize model parameters
\For{each training step $t$}
    \State Sample a candidate batch of $N$ examples from the stream
    \For{each candidate $z$ in the candidate batch}
        \State Score $z$ by how much including $z$ now would reduce validation loss on $V$, evaluated $K$ steps later
    \EndFor
    \State Keep the top-$B$ candidates as the training batch
    \State Update model parameters on the selected training batch
\EndFor
\State \Return trained model
\end{algorithmic}
\end{algorithm}

Two boundary cases are worth highlighting. $K = 0$ recovers the per-step scoring rule used by GREATS~\cite{wang2024greats} (without its greedy selection iteration). The limit of $K$ equal to the remaining training steps is the full-trajectory horizon; it is intractable in the online loop and is instead approximated by offline selection on a completed reference trajectory.

\end{document}